\pdfoutput=1

\documentclass[11pt]{article}

\usepackage[final]{ACL2023}

\usepackage{times}
\usepackage{latexsym}

\usepackage[T1]{fontenc}

\usepackage[utf8]{inputenc}

\usepackage{microtype}

\usepackage{inconsolata}

\usepackage{graphicx}       
\usepackage{multirow}       
\usepackage{amssymb}        
\usepackage{bold-extra}     
\usepackage{bm}             
\usepackage[hang,flushmargin]{footmisc}  
\usepackage{hyperref}
\usepackage{pifont}         
\usepackage{tablefootnote}
\usepackage{tabularx}

\newcommand{\LN}{\linebreak\noindent}   

\newcommand{\cmark}{\ding{51}} 
\newcommand{\xmark}{\ding{55}} 

\usepackage{xparse}
\NewDocumentCommand{\rot}{O{90} O{0.5em} m}{\makebox[#2][l]{\rotatebox{#1}{#3}}} 

\usepackage{float}
\usepackage{enumitem}
\usepackage{booktabs}
\usepackage{xcolor}

\title{Don't Forget Your ABC's: Evaluating the State-of-the-Art in Chat-Oriented Dialogue Systems}

\author{
  Sarah E. Finch\footnotemark[1]
  \and
  James D. Finch\footnotemark[1]
  \and
  Jinho D. Choi
  \\
  Department of Computer Science
  \\
  Emory University
  \\
  Atlanta, GA, USA
  \\
  \texttt{\{sfillwo, jdfinch, jinho.choi\}@emory.edu}
  \\
}

\date{}

\begin{document}
\maketitle

\renewcommand{\thefootnote}{\fnsymbol{footnote}}
\footnotetext[1]{Contributed equally to this work as first authors.}
\renewcommand{\thefootnote}{\arabic{footnote}}

\begin{abstract}

Despite tremendous advancements in dialogue systems, stable evaluation still requires human judgments producing notoriously high-variance metrics due to their inherent subjectivity.
Moreover, methods and labels in dialogue evaluation are not fully standardized, especially for open-domain chats, with a lack of work to compare and assess the validity of those approaches.
The use of inconsistent evaluation can misinform the performance of a dialogue system, which becomes a major hurdle to enhance it.
Thus, a dimensional evaluation of chat-oriented open-domain dialogue systems that reliably measures several aspects of dialogue capabilities is desired.
This paper presents a novel human evaluation method to estimate the rates of many\LN dialogue system behaviors.
Our method is used to evaluate four state-of-the-art open-domain dialogue systems and compared with existing approaches.
The analysis demonstrates that our behavior method is more suitable than alternative Likert-style or comparative approaches for dimensional evaluation of these systems.


\end{abstract}
\section{Introduction}

Recent work in human-computer chat has made remarkable progress. 
Multi-turn open-domain (MTOD) models are capable of holding engaging conversations with humans \cite{roller:21, adiwardana:20}. 
However, there remain a number of challenges facing MTOD chatbots such as hallucinations \cite{shuster:21}, commonsense violations \cite{zhou:21}, and consistency issues \cite{nie:21}. 
A significant obstacle for research that addresses these challenges is the difficulty in formulating an appropriate evaluation methodology due to the inherent subjectivity in determining chat quality \cite{van:21}.  
Since existing automatic evaluation metrics have been shown to be biased measures of chat quality \cite{liu:16, sai:19, deriu:22}, evaluation using human judgments is standard, although the type of human judgments varies widely across works \cite{finch_towards:20}.

Overall, there are few works comparing and assessing the validity of various human evaluation methods.
The result of this gap in the literature is that the relative sensitivity, interpretability, and importance of the metrics used to evaluate chat models are not well understood. 
A dimensional approach for evaluating chat models that measures different aspects of chat quality would surely aid progress \cite{van:21}.
However, to our knowledge, no work has investigated the coverage of a comprehensive set of evaluation metrics. 
Consequently, existing chat model evaluation results provide an incomplete picture of the strengths and weaknesses of MTOD chatbots. 
This paper addresses these limitations of previous work through the following three contributions:

\begin{enumerate}[leftmargin=*]
    \item A novel, dimensional human evaluation method that measures the rate of chatbot behaviors impacting chat quality (Section~\ref{sec:abc-eval}).
    \item A detailed validation of human evaluation methods, including likert scales and pairwise comparisons (Section~\ref{sec:metric-analysis}).
    \item A comprehensive evaluation of four MTOD chatbots using validated metrics (Section~\ref{sec:bot_evaluation_results}).
\end{enumerate}

\noindent By presenting a detailed picture of MTOD chatbot performance and standard methods to evaluate them, we aid future work’s efforts to further understand and improve human-computer interaction. Our evaluation platform, analyses, and data are available at \url{https://github.com/emorynlp/ChatEvaluationPlatform}. 

\section{Chatbots}

To evaluate the strengths and weaknesses of MTOD models, we select the chatbots for our study using a two-stage process: (1) a literature review to identify chatbot candidates, and (2) a pilot evaluation to select the final set of bots for our full study.

\paragraph{Literature Review} To promote diversity among the selected chatbots, we focus our review on four popular themes of the human-computer chat: (1) Knowledge-grounded chat, (2) Empathetic chat, (3) Self-consistent chat, and (4) General open-domain chat with large pre-training resources like Reddit. 
Candidate chatbots are selected from each theme using the following criteria:

\begin{enumerate}[leftmargin=*]
\setlength\itemsep{0em}
    \item The bot must demonstrate state-of-the-art performance in a task related to the theme.\footnote{Note that selection occurred in October 2021.}
    \item The implementation must be provided.\footnote{We accepted either a trained English model or codebase with a fully-specified procedure to replicate the model.}
    \item The response latency of the bot must be <10 seconds using modern GPU hardware.
\end{enumerate}

\noindent This review yields the 6 chatbot candidates in Table \ref{tab:bot_pilot_results}: Blender-Decode \cite{nie:21}, Blender2 \cite{weston:21}, BART-FiD-RAG \cite{shuster:21}, Emora \cite{finch_emora:20}, DukeNet \cite{meng:20}, and CEM \cite{sabour:22}. Appendix~\ref{sec:chatbot_selection} presents details of our literature review and selection process. 

\begin{table}[htbp]
\resizebox{\columnwidth}{!}{
\begin{tabular}{l|c|r|c|c}
    \toprule
    \multicolumn{1}{c|}{\textbf{Model}} & \textbf{Theme} & \textbf{N} & \textbf{Q} & \textbf{Pass} \\ 
    \midrule
    Blender-Decode                      &  Consistency   &         10 &    4.1     &    \cmark     \\
    Blender2                            &    General     &         10 &    3.8     &    \cmark     \\
    BART-FiD-RAG                         &   Knowledge    &         10 &    3.5     &    \cmark     \\
    Emora                               &    General     &         10 &    3.3     &    \cmark     \\
    DukeNet                             &   Knowledge    &          9 &    1.9     &    \xmark     \\
    CEM                                 &    Empathy     &         12 &    1.1     &    \xmark \\
    \bottomrule
\end{tabular}%
}

\caption{The pilot results for 6 bots, showing the theme of the approach (\textbf{Theme}), the number of collected conversations (\textbf{N}), and the avg.\ dialogue-level Likert quality rating (\textbf{Q}). \textbf{Pass} denotes which models passed the verification criteria and were included in the full study.}
\label{tab:bot_pilot_results}
\vspace{-2ex}
\end{table}

\paragraph{Chatbot Selection}
A pilot evaluation using the 6 chatbot candidates is conducted in order to verify the multi-turn dialogue capability of the chatbot candidates. Appendix \ref{sec:chatbot_implementations} provides details on the implementations of each chatbot candidate. 10 students majoring in Computer Science or Linguistics are invited to interact with randomly assigned chatbots in 3-5 text-based conversations,\footnote{We use the web interface provided by ParlAI \cite{miller:17} hosted on our local webserver.} each of which consisted of 30 turns.\footnote{A ``turn'' is defined as ONE message from a single interactor.}
At the end of each conversation, students are asked to rate the quality from 1 (least) to 5 (most). Based on the pilot results (Table \ref{tab:bot_pilot_results}), DukeNet and CEM are excluded from our full study because they are unable to hold satisfying multi-turn conversations, despite their reasonable single-turn response generation capabilities. Appendix~\ref{sec:bot_pilot_examples} shows example dialogues from these systems.


\section{Conversation Collection}
\label{sec:conversations}

The conversation dataset used for the full study is collected using human interactors in a text-based conversation setting. 
46 undergraduates are recruited as interactors.
Each interactor is compensated with a \$5 gift card for every 6 conversations, and allowed to complete up to 18 conversations. 
Conversations are collected remotely using ParlAI’s interactive web interface, and links to the web interface are sent to each interactor with instructions to be completed within 2 weeks.

For each link, the interactor completes two conversations with a random pair of chatbots, for a minimum of 30 turns per conversation. 
We impose a similar open-ended, topic-free chatting environment to \citet{adiwardana:20}.
Interactors are asked to rate 8 dimensions (Table \ref{tab:dims}) of each conversation after its completion on a 1-5 Likert scale, and to select the higher-quality conversation along the same 8 dimensions after each conversation-pair (ties allowed). Our final conversation dataset includes 400 human-bot dialogues (100 dialogues per chatbot), averaging 30.3 turns per dialogue (11.3 tokens per user turn).

\section{Evaluation Methods}
\label{sec:eval_prev_works}


For a comprehensive evaluation of MOTD chatbots, a robust dimensional evaluation of their chat capabilities is crucial \cite{van:21}. 
To have confidence that any evaluation metric yields useful information, its interpretability and sensitivity require validation. In addition, it is important to verify that each evaluation metric provides distinct information relative to the others.

Several previous works propose sets of evaluation metrics that could be used for a dimensional evaluation but with insufficient analyses to validate them. 
\citet{finch_towards:20} present an exhaustive set of metrics based on a literature survey of human evaluation methods, but do not quantitatively validate its interpretability, sensitivity, or per-metric distinctness. 
\citet{mehri_unsupervised:20} present a set of Likert metrics and analyze their relationship to overall dialogue quality, but do not validate the sensitivity or distinctness of the individual metrics. 
\citet{mehri_usr:20} present 5 Likert metrics and evaluate their coverage with respect to explaining single response quality, but do not validate their sensitivity or distinctness. 

Similarly, some works look to identify common chatbot errors. \citet{sanguinetti:20} and \citet{higashinaka:21} present error taxonomies empirically grounded by error analyses, but do not present distinctness or sensitivity results for their error categories. 
\citet{see:21} identify errors for one dialogue model and analyze the impact of each error on overall quality but do not attempt to verify the generalizability of their results. 

Furthermore, various works propose novel evaluation methods with varying degrees of validation of the reliability and effectiveness of such methods. \citet{deriu:20} present Spot the Bot, a pairwise evaluation approach that uses survival analysis to rank bots based on self-chats, but do not directly compare to alternative methodologies other than for cost. \citet{sedoc:20} apply Item-Response Theory (IRT) \cite{lord:08} to pairwise comparison dialogue evaluation, by using a latent variable Bayesian model to estimate both the ability of the evaluated systems and the informativeness of inputs in the static evaluation set. Their analysis of the utility of IRT for dialogue evaluation does not include comparisons to existing approaches or a dimensional focus since they exclusively consider overall response quality. \citet{ji:22} propose a continuous-scale method for evaluating multi-turn dialogue systems with quality control measures for mitigating artifacts from human annotators. They validate their proposed method on various dialogue dimensions using replication studies, a sensitivity analysis, and a correlation analysis between dimensions, although they explicitly acknowledge that their set of dimensions is not intended to be comprehensive. \citet{phy:20} assert 3 dimensions (understandability, sensibleness, and likability) are sufficient for capturing the quality of a dialogue and validate their claims using agreement, correlation analysis, and distinctness analysis on human annotations of their dimensions, although they are not applied to multi-turn dialogues.

\noindent Two studies, \citet{li_acute-eval:19} and \citet{smith:22}, compare pairwise comparison and Likert evaluation methods via a sensitivity analysis. However, neither of them target a high-coverage set of dimensional metrics, as their studies were limited to 4 and 3 metrics respectively. \citet{lee:20} also investigates pairwise evaluation using the ChatEval platform. However, this is not a multi-turn evaluation setup and it does not target a dimensional analysis since the comparisons are based exclusively on the overall quality of the responses. 

\begin{table}[htp!]
\centering
\resizebox{\columnwidth}{!}{%
\begin{tabular}{l|ccccccc}
\toprule
 & \bf M & \bf C & \bf P & \bf A & \bf S & \bf I & \bf D\\
\midrule
\citet{finch_towards:20}   & \cmark & \cmark &  &  &  & \cmark &  \\
\citet{mehri_unsupervised:20}   & \cmark & \cmark & \cmark & \cmark &  & \cmark &  \\
\citet{mehri_usr:20}   &  & \cmark &  & \cmark &  & \cmark &  \\
\citet{sanguinetti:20}   & \cmark & \cmark &  & \cmark &  &  &  \\
\citet{higashinaka:21}   & \cmark & \cmark & \cmark & \cmark &  &  &  \\
\citet{see:21}  & \cmark &  &  & \cmark &  & \cmark &  \\
\citet{deriu:20}  & \cmark &  &  & \cmark & \cmark & \cmark &  \\
\citet{sedoc:20}  &  &  &  & \cmark & \cmark &  &  \\
\citet{ji:22}   & \cmark &  &  & \cmark & \cmark & \cmark &  \\
\citet{phy:20}   &  &  &  & \cmark &  & \cmark & \cmark \\
\citet{li_acute-eval:19} & \cmark &  & \cmark &  & \cmark &  &  \\
\citet{smith:22}   & \cmark &  & \cmark &  & \cmark &  &  \\
\citet{lee:20} &  &  &  & \cmark & \cmark &  &  \\
\midrule
\bf This Work &  \cmark & \cmark & \cmark & \cmark & \cmark & \cmark & \cmark \\
\bottomrule
\end{tabular}%
}
\caption{Recent studies of human evaluation metrics, in order of mention in Section \ref{sec:eval_prev_works}.
\textbf{M}ulti-turn: investigates multi-turn response generation,
\textbf{C}omprehensive: a set of metrics intended to explain dialogue quality,
Com\textbf{P}ared: compares alternative evaluation methods,
\textbf{A}greement: inter-annotator agreement,
\textbf{S}ensitivity: validates metric sensitivity via statistical testing,
\textbf{I}mportance: relates evaluation metrics to overall dialogue quality,
\textbf{D}istinctness: analyzes whether metrics provide distinct information about quality.
}
\label{tab:existing_eval_methods}
\vspace{-1.2ex}
\end{table}


\noindent Overall, the relative validity of human evaluation metrics requires further investigation before a comprehensive and reliable dimensional evaluation of human-computer chat is achieved. Table~\ref{tab:existing_eval_methods} summarizes the goals and contributions of the previous evaluation works. 
Our study addresses all existing gaps by conducting a detailed validation study of 4 different human evaluation methods and a wide range of fine-grained metrics.

\subsection{Selected Methods}
\label{sec:selected_evals}

Four human evaluation methods are chosen for our study. 
Since MTOD chat model evaluation is our goal, any domain- or approach-specific methods or single-response evaluation methods providing chatbots with a specific context are excluded.\footnote{We do not include a turn-level comparative evaluation because controlled comparisons require comparing turns with identical historical contexts which is not viable for real human-bot dialogues like those used in this work.}
We also focus on external human evaluation methods, where human evaluators judge conversations they do not participate in. 
Three of the selected methods represent popular approaches: Dialogue Likert, Turn Likert, and Comparative. 
The fourth method, ABC-Eval, is our novel evaluation approach. 



\begin{table}[htbp!]
\centering
\resizebox{\columnwidth}{!}{
\begin{tabular}{l|c|c|c}
\toprule
\multicolumn{1}{c|}{\multirow{2}{*}{\bf Label}} & \bf Dialogue & \bf Turn & \multirow{2}{*}{\bf Comparative} \\
& \bf Likert & \bf Likert & \\
\midrule
Consistency & \textcolor{red}{\tt Con$_d$} & \textcolor{blue}{\tt Con$_t$} & \textcolor{black!50!green}{\tt Con$_c$} \\
Emotion & \multirow{2}{*}{\textcolor{red}{\tt Emo$_d$}} & \multirow{2}{*}{\textcolor{blue}{\tt Emo$_t$}} & \multirow{2}{*}{\textcolor{black!50!green}{\tt Emo$_c$}}\\
Understanding & & & \\
Engagingness & \textcolor{red}{\tt Eng$_d$} & \textcolor{blue}{\tt Eng$_t$} & \textcolor{black!50!green}{\tt Eng$_c$}\\
Grammaticality & \textcolor{red}{\tt Gra$_d$} & \textcolor{blue}{\tt Gra$_t$} & \textcolor{black!50!green}{\tt Gra$_c$}\\
Informativeness & \textcolor{red}{\tt Inf$_d$} & \textcolor{blue}{\tt Inf$_t$} & \textcolor{black!50!green}{\tt Inf$_c$}\\
Quality & \textcolor{red}{\tt Qua$_d$} & \textcolor{blue}{\tt Qua$_t$} & \textcolor{black!50!green}{\tt Qua$_c$}\\
Proactivity & \textcolor{red}{\tt Pro$_d$} & \textcolor{blue}{\tt Pro$_t$} & \textcolor{black!50!green}{\tt Pro$_c$}\\
Relevance & \textcolor{red}{\tt Rel$_d$} & \textcolor{blue}{\tt Rel$_t$} & \textcolor{black!50!green}{\tt Rel$_c$}\\
\bottomrule
\end{tabular}
}
\caption{The 8 labels for Likert and Comparative evaluations (taken from \citet{finch_towards:20}), henceforth referred to using their abbreviations and colors.}
\label{tab:dims}
\vspace{-2ex}
\end{table}

\begin{table*}[ht]
\resizebox{\textwidth}{!}{%
\begin{tabular}{l|r|l|c}
\toprule
\multicolumn{1}{c|}{\textbf{Label}} & \multicolumn{1}{c|}{\textbf{Abbr.}} & \multicolumn{1}{c|}{\textbf{Description}} & \multicolumn{1}{c}{\textbf{Inspired by}} \\
\midrule
Uninterpretable & \textcolor{orange!80!red}{\textbf{\texttt{!Int$_b$}}} & It is difficult to understand the intended meaning of part or all of the response. & 1, 2, 3, 4, 5, 6 \\
\midrule
Antisocial & \textcolor{orange!80!red}{\textbf{\texttt{!Soc$_b$}}} & The response is insulting, hateful, or excessively vulgar. & 2, 7, 8, 9  \\ 
\midrule
Preference Info & \textcolor{orange!80!red}{\textbf{\texttt{Pre$_b$}}} & The response expresses the bot's preferences, wishes, or values. & \multirow{2}{*}{10, 11} \\ 
Life Info & \textcolor{orange!80!red}{\textbf{\texttt{Lif$_b$}}} & The response shares information about the bot's life or experiences. & \\
\midrule
\bf Empathetic & \textcolor{orange!80!red}{\textbf{\texttt{Emp$_b$}}} & The response shows an understanding and reacts appropriately to someone's emotions. &  \multirow{2}{*}{11, 12, 13} \\
\bf Lack of Empathy & \textcolor{orange!80!red}{\textbf{\texttt{!Emp$_b$}}} & The bot misunderstands or reacts inappropriately to someone's emotions. &  \\
\midrule
\bf Commonsense & \multirow{2}{*}{\textcolor{orange!80!red}{\textbf{\texttt{!Com$_b$}}}} & \multirow{2}{*}{The response misunderstands or contradicts common knowledge.} & \multirow{2}{*}{2, 14, 15, 16}  \\ 
\bf Contradiction & & & \\
\midrule
Fact Usage & \textcolor{orange!80!red}{\textbf{\texttt{Fac$_b$}}} & The response accurately incorporates encyclopedic or expert knowledge. & 1, 2, 11, 17, 18,\\ 
\bf Fact Contradiction & \textcolor{orange!80!red}{\textbf{\texttt{!Fac$_b$}}} & The response hallucinates or inaccurately presents encyclopedic or expert knowledge. &  19, 20 \\ 
\midrule
\bf Self Contradiction & \textcolor{orange!80!red}{\textbf{\texttt{!Sel$_b$}}} & The bot contradicts something it said earlier in the dialogue. & \multirow{2}{*}{2, 3, 6, 20, 21,}   \\ 
\bf Partner Contradiction & \textcolor{orange!80!red}{\textbf{\texttt{!Par$_b$}}} & The bot contradicts or misremembers something the user said earlier in the dialogue. &  \\
\bf Redundant & \textcolor{orange!80!red}{\textbf{\texttt{Red$_b$}}} & The response inappropriately repeats information presented earlier in the dialogue. & \multirow{-2}{*}{22, 23} \\
\midrule
\bf Ignore & \textcolor{orange!80!red}{\textbf{\texttt{Ign$_b$}}} & The response ignores what the user just said. &  \multirow{4}{*}{1, 2, 3, 6, 24}   \\ 
\bf Irrelevant & \textcolor{orange!80!red}{\textbf{\texttt{!Rel$_b$}}} & The response interrupts the current topic of discussion by presenting unrelated information. & \\
Follow-up & \textcolor{orange!80!red}{\textbf{\texttt{Fol$_b$}}} & The response explores, elaborates on, or asks about the ideas shared in the previous turn. & \\
Topic Switch & \textcolor{orange!80!red}{\textbf{\texttt{Top$_b$}}} & The response introduces a new topic of conversation. & \\
\bottomrule
\end{tabular}
}
\caption{The 16 behavior labels within ABC-Eval. Row separators denote evaluation task groupings. \textbf{Bold} indicates behavior labels kept in final set. 
[1] \citet{gopalakrishnan:19}, 
[2] \citet{higashinaka:21},
[3] \citet{mehri_unsupervised:20}, 
[4] \citet{mehri_usr:20}, 
[5] \citet{phy:20},  
[6] \citet{sanguinetti:20}, 
[7] \citet{beattie:22}, 
[8] \citet{sun:22},
[9] \citet{xu:21}, 
[10] \citet{rashkin:21}, 
[11] \citet{smith:20}, 
[12] \citet{majumder:20},
[13] \citet{rashkin:19}, 
[14] \citet{zhong:21}, 
[15] \citet{zhou:21}, 
[16] \citet{zhou:22},
[17] \citet{gupta:22}, 
[18] \citet{honovich:21}, 
[19] \citet{santhanam:21}, 
[20] \citet{shuster:21}, 
[21] \citet{li:21}, 
[22] \citet{nie:21}, 
[23] \citet{welleck:19}, 
[24] \citet{xu:22} 
.}
\label{tab:behavior_labels}
\vspace{-3ex}
\end{table*}

\paragraph{Dialogue Likert} 
\label{sec:dialogue-likert}
Annotators provide dialogue-level ratings from 1 (least) to 5 (most) for the 8 labels shown in Table \ref{tab:dims}. 
We use the dimension set proposed in \citet{finch_towards:20} which results from a detailed survey of characteristics used in chat evaluation and has better coverage than alternatives like the set used in ACUTE-Eval \cite{li_acute-eval:19}. 
Bot-level metrics are calculated as the mean rating across all bot dialogues.

\paragraph{Turn Likert}
Annotators provide turn-level ratings on the same scale and labels as those used for Dialogue Likert. 
The dialogue-level metric is measured as the mean rating of a single dialogue's turns. 
The bot-level metric is calculated as the mean rating of all turns in all bot dialogues.

\paragraph{Comparative}
Annotators select the dialogue in which chatbot responses better fit a label definition from a side-by-side pair of dialogues, also using the labels in Table \ref{tab:dims}. 
A “neither” option is allowed, only for cases where the evaluator cannot distinguish which dialogue was a better fit. 
Bot-level metrics are calculated as bot pair win/tie/loss proportions between pairing of their dialogues.

\paragraph{Behavior Classification: ABC-Eval}
Annotators provide binary labels on the turn-level indicating the presence or absence of a particular chat characteristic. The included chat characteristics are defined in Table \ref{tab:behavior_labels}.
Dialogue-level metrics are calculated as the proportion of turns that display the characteristic of the dialogue. 
Bot-level metrics are calculated as the proportions of turns that display the characteristic over all bot dialogues.
ABC-Eval is described in detail next in Section~\ref{sec:abc-eval}.

\section{ABC-Eval Design}
\label{sec:abc-eval}

We hypothesize that binary turn-level behavior labels provide more reliable and informative metrics for quantifying fine-grained aspects of chat quality than alternative approaches such as Likert or Comparative scoring. 
Our novel method, the Annotation of Behaviors in Chat Evaluation (ABC-Eval), is developed in three stages: (1) collecting a set of behavior label candidates, (2) developing and piloting our annotation instructions and procedure, and (3) selecting a subset of behavior labels based on the validation study results in Section~\ref{sec:metric-analysis}.

\paragraph{Collecting Behavior Label Candidates}
\label{sec:label_cands}
Based on a review of recent work in chat-oriented dialogue modeling and evaluation, we identify characteristics of chatbot responses relevant to conversation quality. 
These characteristics include those presented as error cases, evaluation metrics, or desirable response features. 
We then curate binarized definitions of these characteristics to create an initial set of behavior label candidates, which are revised through an iterative piloting and development process. 
Due to its high coverage of error categories, \citet{higashinaka:21} is the primary source of inspiration for many of our behavior labels.
However, we improve upon their presented taxonomies by considering additional labels based on characteristics of chat presented by other work, and by further refining their error categories to improve average Inter-Annotator Agreement (Section~\ref{sec:agreements}). 
Table \ref{tab:behavior_labels} presents the final set and definitions of the 16 candidate behavior labels used in our full study, along with selected works from our review that inspired their inclusion.
Appendix~\ref{sec:label_pilots} details in full our development process.

\paragraph{Annotation Procedure}
The ABC-Eval procedure includes 16 binary behavior labels divided between 8 independent annotation tasks (Table \ref{tab:behavior_labels}). 
In each task, human evaluators are provided with definitions and examples of the behavior labels associated with that task and asked to annotate every chatbot turn in a given human-chatbot conversation with each behavior label. 
Evaluators complete these tasks using a custom web application based on the ParlAI evaluation interface (Appendix~\ref{sec:interfaces}). 




\vspace{-0.5ex}
\paragraph{Training and Screening}
To improve annotation consistency and detect poorly performing evaluators, we develop automated training sessions each annotation task inspired by \citet{van:21}. 
Each session consists of 3 conversations that evaluators annotate using an identical procedure and web interface to the corresponding task. 
The 3 conversations used for each session are hand-crafted by the authors to represent a variety of positive and negative examples of the behavior labels for the corresponding task (Appendix \ref{sec:label_pilots}). 
The gold annotations for each training conversation are hidden from evaluators during the annotation; however, after completing each training conversation, any disagreements between the evaluator’s annotations and gold labels are displayed along with an explanation to help the evaluator improve. 
We use the evaluator’s performance on the third conversation of each training session to screen evaluators, where performance is measured by the number of turns where their annotations disagree with gold labels. 
Evaluators are eligible to complete the work on a task if they make mistakes on fewer than 2 turns for the antisociality and uninterpretability tasks, or on fewer than 3 turns for the other 6 tasks.

\addtocounter{footnote}{3}

\begin{figure*}[!ht]
\centering
\includegraphics[width=\textwidth]{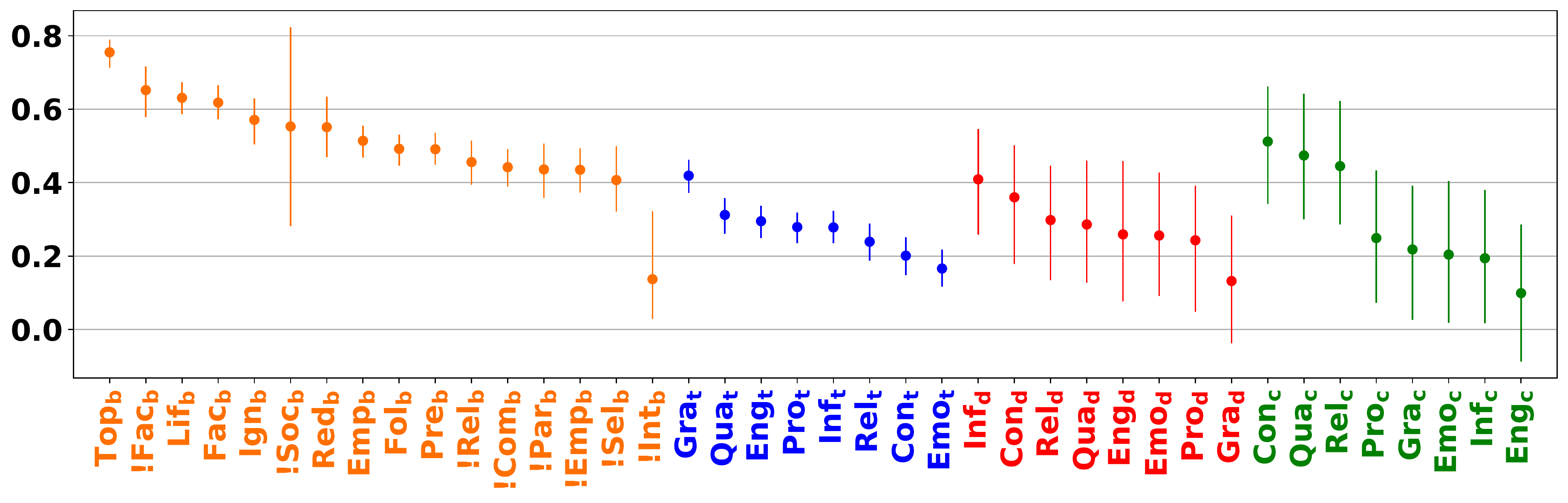}
\caption{IAA (Krippendorff's alpha) for all metrics. Error bars denote 95\% bootstrap\footnotemark confidence intervals.\footnotemark}
\label{fig:surge_agreement}
\vspace{-2ex}
\end{figure*}

\addtocounter{footnote}{-3}

\section{Evaluation Study}
\label{sec:eval_study}

Our full study consists of the collection of 40 labels per conversation. This collection was split into 18 independent evaluation tasks as follows:

\begin{itemize}
    \item 8 ABC-Eval tasks, each composed of 1 to 4 labels as denoted by groupings in Table \ref{tab:behavior_labels}
    \item 1 Dialogue Likert task, composed of all 8 labels from Table \ref{tab:dims} completed in random order
    \item 8 Turn Likert tasks, each composed of 1 label from Table \ref{tab:dims}
    \item 1 Comparative task, composed of all 8 labels from Table \ref{tab:dims} completed in random order
\end{itemize}

\addtocounter{footnote}{-2}

\noindent The 18 evaluation tasks are posted on SurgeHQ’s annotation platform\footnote{\href{https://www.surgehq.ai}{https://www.surgehq.ai}; Appx. \ref{sec:worker_analysis} details annotator selection.} to be completed by dedicated remote workers (Surgers) with experience in NLP annotation. 
Each time an evaluator connects to one of our tasks, they are assigned a randomly selected conversation to annotate. 
We are allocated a group of 125 Surgers, chosen by a SurgeHQ employee based on high annotation performance on past projects.
Evaluators are compensated per annotated conversation per task, at an estimated rate of \$20/hr\footnote{Per-task payment rates provided in Appendix \ref{sec:collection_cost}.}. 
We allow evaluators to annotate up to 60 conversations per task.

\noindent Our final evaluation dataset consists of 400 conversations, each with results for all 40 labels.\footnote{Only 192 of our 200 dialogue pairs were evaluated with Comparative labels due to a collection mistake}
Additionally, a randomly-selected subset of 100 conversations (and 50 of the conversation pairs) is evaluated a second time by a different Surger in order to measure IAA. 

\addtocounter{footnote}{1}
\footnotetext{Bias-corrected and accelerated confidence intervals with $k=$10,000 Monte Carlo case resamples.}
\stepcounter{footnote}
\footnotetext{$!Soc_b$ and $!Int_b$'s confidence intervals are largely due to a low rate of positive examples (see Figure \ref{fig:undesirable_behaviors}).}

\section{Metric Analysis}
\label{sec:metric-analysis}

\subsection{Interpretability}
\label{sec:agreements}

We measure the reliability of interpreting each metric's annotation instructions by calculating IAA using our set of 100 double-annotated conversations (Figure~\ref{fig:surge_agreement}). 
High agreement between annotators demonstrates that different people can reliably come to the same conclusions about how a metric's definition applies to each chatbot response.

Our results suggest that the definitions of most ABC-Eval metrics can be interpreted more reliably than the definitions of most Dialogue Likert, Turn Likert, and Dialogue Comparison metrics. Likert-style and comparison-style annotations appear to have similar interpretability, although $Qua_c$ was a notable exception that produced higher agreement than $Qua_d$.

\subsection{Importance}
\label{sec:importance}

\begin{figure*}[ht]
\centering
\includegraphics[width=\textwidth]{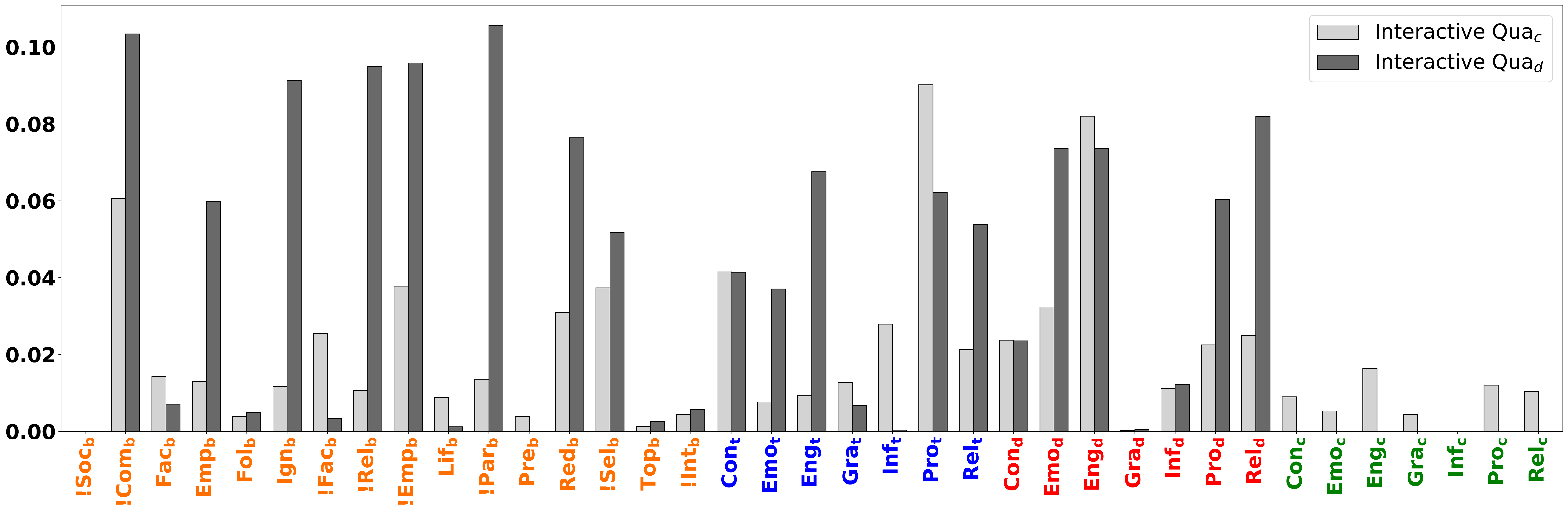}
\caption{Extent to which each evaluation metric can explain variance in conversation quality by fitting a univariate regression model ($R^2$ for predicting $Qua_d$ using linear regression, McFadden’s Pseudo-$R^2$ for predicting $Qua_c$ using logistic regression). Comparative metrics cannot predict $Qua_d$ so only results for variance of $Qua_c$ are shown.}
\label{fig:predictive_validity}
\end{figure*}

\begin{table*}[ht]
\centering
\resizebox{\textwidth}{!}{%
\begin{tabular}{l | c c c c c c c c c c c c c c c c | c c c c c c c c | c c c c c c c c | c c c c c c c c }

$\alpha$ & \rot{!Soc$_b$} & \rot{!Com$_b$} & \rot{Fac$_b$} & \rot{Emp$_b$} & \rot{Fol$_b$} & \rot{Ign$_b$} & \rot{!Fac$_b$} & \rot{!Rel$_b$} & \rot{!Emp$_b$} & \rot{Lif$_b$} & \rot{!Par$_b$} & \rot{Pre$_b$} & \rot{Red$_b$} & \rot{!Sel$_b$} & \rot{Top$_b$} & \rot{!Int$_b$} & \rot{Con$_t$} & \rot{Emo$_t$} & \rot{Eng$_t$} & \rot{Gra$_t$} & \rot{Inf$_t$} & \rot{Pro$_t$} & \rot{Qua$_t$} & \rot{Rel$_t$}  & \rot{Con$_d$} & \rot{Emo$_d$} & \rot{Eng$_d$} & \rot{Gra$_d$} & \rot{Inf$_d$} & \rot{Pro$_d$} & \rot{Qua$_d$} & \rot{Rel$_d$} & \rot{Con$_c$} & \rot{Emo$_c$} & \rot{Eng$_c$} & \rot{Gra$_c$} & \rot{Inf$_c$} & \rot{Pro$_c$} & \rot{Qua$_c$} & \rot{Rel$_c$}  \\
\toprule
0.01 & 0 & 1 & 4 & 3 & 5 & 1 & 4 & 2 & 4 & 2 & 0 & 4 & 2 & 5 & 5 & 2 & 1 & 1 & 2 & 4 & 4 & 4 & 4 & 2 & 1 & 2 & 1 & 0 & 0 & 3 & 1 & 0 & 1 & 0 & 1 & 0 & 1 & 1 & 0 & 0 \\
0.05 & 0 & 2 & 5 & 3 & 5 & 2 & 6 & 2 & 5 & 2 & 1 & 4 & 3 & 5 & 5 & 3 & 2 & 1 & 3 & 5 & 4 & 4 & 4 & 3 & 2 & 3 & 2 & 3 & 2 & 3 & 1 & 2 & 1 & 0 & 1 & 0 & 1 & 2 & 1 & 0 \\
0.1 & 1 & 3 & 5 & 3 & 5 & 2 & 6 & 2 & 5 & 3 & 1 & 4 & 3 & 5 & 5 & 3 & 2 & 3 & 3 & 5 & 4 & 4 & 4 & 4 & 2 & 4 & 2 & 5 & 3 & 3 & 3 & 2 & 2 & 0 & 1 & 0 & 2 & 2 & 1 & 2
\end{tabular}%
}
\caption{The number of statistically significant differences detected by each metric when comparing bot-pairs using z-tests of proportions (ABC-Eval), t-tests (Turn Likert and Dialogue Likert), and sign tests (Comparative) at three significance thresholds.}
\label{tab:sensivity}
\vspace{-2ex}
\end{table*}

The importance of each metric is estimated by a predictive validity analysis that measures the extent, to which the metric can predict conversation quality (Figure~\ref{fig:predictive_validity}). 
We use $Qua_d$ and $Qua_c$ from interactors that participated in the conversations (Section~\ref{sec:conversations}) to avoid cases where the same evaluator produced the quality label and explanatory metric. 
The predictive validity of each metric was measured by fitting univariate linear or logistic regression models to predict $Qua_d$ or $Qua_c$, respectively. 

$Qua_c$ was represented as a binary encoding, where 0 and 1 represent choosing the first and second conversation, respectively. We excluded any conversation pairs in which the interactor could not distinguish a difference in quality between conversations, and fitted models on the remaining set of 184 conversations. To use non-comparative predictors for predicting $Qua_c$, the difference in metric value between each pair of conversations was used.

Our results suggest that dialogue quality is substantially related to emotional understanding metrics ($Emo$, $Emp_b$, $!Emp_b$), relevance-related metrics ($Rel$, $!Rel_b$, $Ign_b$), and consistency metrics ($Con$, $!Sel_b$, $Red_b$, $!Par_b$). Within these metric groupings, ABC-Eval metrics were overall more predictive of quality than their Likert or comparative analogs, while comparative metrics were least predictive of quality. Chatbots' ability to express knowledge ($Inf$, $Fac_b$, $!Fac_b$ $Lif_b$, $Pref_b$) was an overall poor predictor of quality; however, commonsense knowledge errors ($!Com_b$) was one of the strongest predictors.

\subsection{Sensitivity}
\label{sec:sensitivity}

We investigate the sensitivity of each metric using two analyses. First, we use the fitness of the univariate regression models described in the previous section as one source of evidence for metric sensitivity, since a metric must be sufficiently sensitive in order to distinguish conversations of low and high quality. Second, we follow \citet{li_acute-eval:19} and run hypothesis tests to count the number of statistically significant differences each metric is able to detect between the 6 pairings of our 4 chatbots (Table~\ref{tab:sensivity}). To make results comparable, we downsample the conversations used for hypothesis testing to 32 conversations per bot for our Dialogue Likert, Turn Likert, and ABC-Eval metrics to match the 32 conversation-pairs per bot-pair produced by our Comparative evaluation.


Our results show that the Likert evaluations were more sensitive than the Comparative evaluation for most labels. ABC-Eval metrics have a wide range of sensitivity, with some ABC-Eval metrics appearing to be more sensitive analogs of similar likert metrics. For example, the results suggest that $!Sel_b$ and $Red_b$ are more sensitive than $Con$, that $Fac_b$ and $!Fac_b$ are more sensitive than $Inf$, and that $Emp_b$ and $!Emp_b$ are more sensitive than $Emo$. On the other hand, the likert-style $Rel$ metric shows similar or slightly superior sensitivity compared to the analogous $Ign$ and $!Rel$ behavior metrics.


\begin{figure*}[ht!]
\centering
\includegraphics[width=\textwidth]{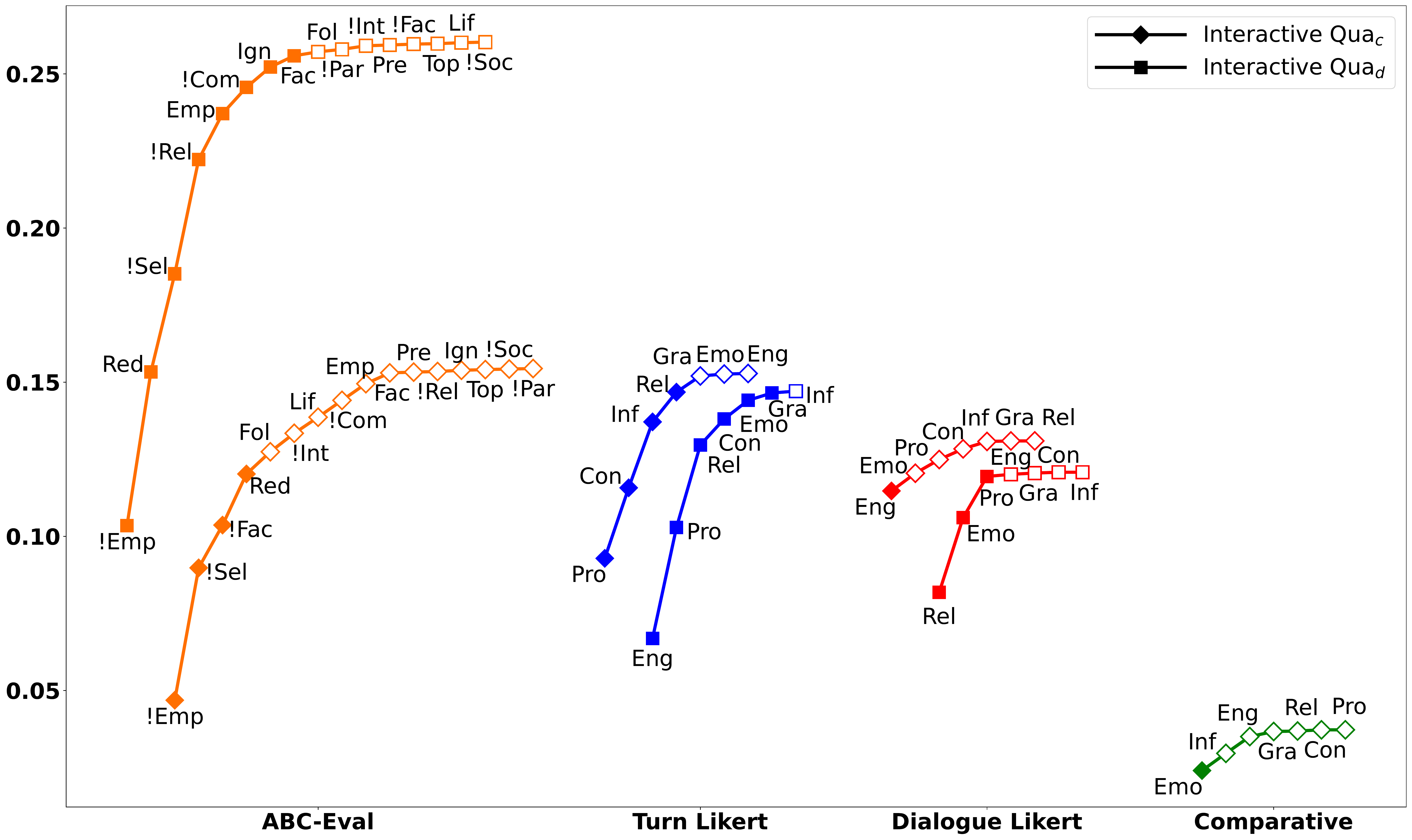}
\caption{Incremental validity of metrics within 4 evaluation methods, obtained using backwards stepwise regression. Points represent the extent to which a model can explain variance in quality ($R^2$ for predicting $Qua_d$ with a linear model, McFadden's pseudo-$R^2$ for predicting $Qua_c$ with a logistic model) using all metrics on the same line and to the left as predictors. Filled marker symbols denote steps where the model’s predictors all contributed positively to adjusted $R^2$ or adjusted pseudo-$R^2$ values; otherwise, marker symbol is unfilled. Comparative metrics cannot be used to predict $Qua_d$ so only results for explaining variance of $Qua_c$ are shown.}
\label{fig:incremental_validity}
\vspace{-1ex}
\end{figure*}

\subsection{Coverage \& Distinctness}
\label{sec:distinctness}

We investigate the coverage and distinctness of our metrics via incremental validity analysis. For this analysis, we perform backwards stepwise regression that determines (1) the ability of an evaluation method as a whole to explain conversation quality, and (2) whether each metric contributes distinct information about quality above and beyond other metrics (Figure~\ref{fig:incremental_validity}). Specifically, we fit a multivariate regression model for each of our 4 evaluation methods. These models are fit similarly to those presented in Section~\ref{sec:importance}, but include all non-quality metrics within an evaluation method as predictors. Then, we remove predictors from each model one at a time based on a beam search ($k$=100) of which removed predictor results in the smallest decrease in model fitness (adjusted $R^2$ or adjusted pseudo-$R^2$). We perform this stepwise regression analysis twice to predict both $Qua_d$ and $Qua_c$ given by interactors, similar to our analysis in Section~\ref{sec:importance}.

\noindent Our results suggest that ABC-Eval has overall better coverage than other evaluation methods for explaining conversation quality. Furthermore, most ABC-Eval metrics that have a strong relationship with conversation quality appear to be appropriately distinct in the information they provide, especially $!Emp_b$, $!Sel_b$, $Red_b$, $!Rel_b$, $Emp_b$, $!Com_b$, and $Ign_b$. Similar distinctness can also be seen in Turn Likert metrics, whereas dialogue-level metrics show relatively low distinctness. 


\subsection{Final ABC-Eval Metrics}

Given the results of our metric analysis, we select the final set of ABC-Eval metrics bolded in Table \ref{tab:behavior_labels}. In our analyses, this final set had better interpretability (Section~\ref{sec:agreements}), a wider coverage of distinct characteristics of chat that impact quality (Section~\ref{sec:importance} and Section~\ref{sec:distinctness}), and overall higher measurement sensitivity (Section~\ref{sec:sensitivity}) than alternative evaluation methods. Furthermore, the final ABC-Eval metrics are less costly\footnote{See Appendix \ref{sec:collection_cost} for detailed cost results.} (a median of 15.2 min/dialogue) to collect than Turn Likert metrics (19.9 min/dialogue). Although dialogue-level evaluations are least costly (2.8 min/dialogue for Dialogue Likert, 4.4 min/dialogue for Comparative), our results suggest that dialogue-level judgements may be ill-suited for dimensional evaluation, since the dialogue-level metrics we tested had worse coverage and distinctness (Section~\ref{sec:distinctness}).


\section{Chatbot Evaluation}
\label{sec:bot_evaluation_results}

\begin{figure*}[ht!]
    \centering
    \includegraphics[width=0.9\textwidth]{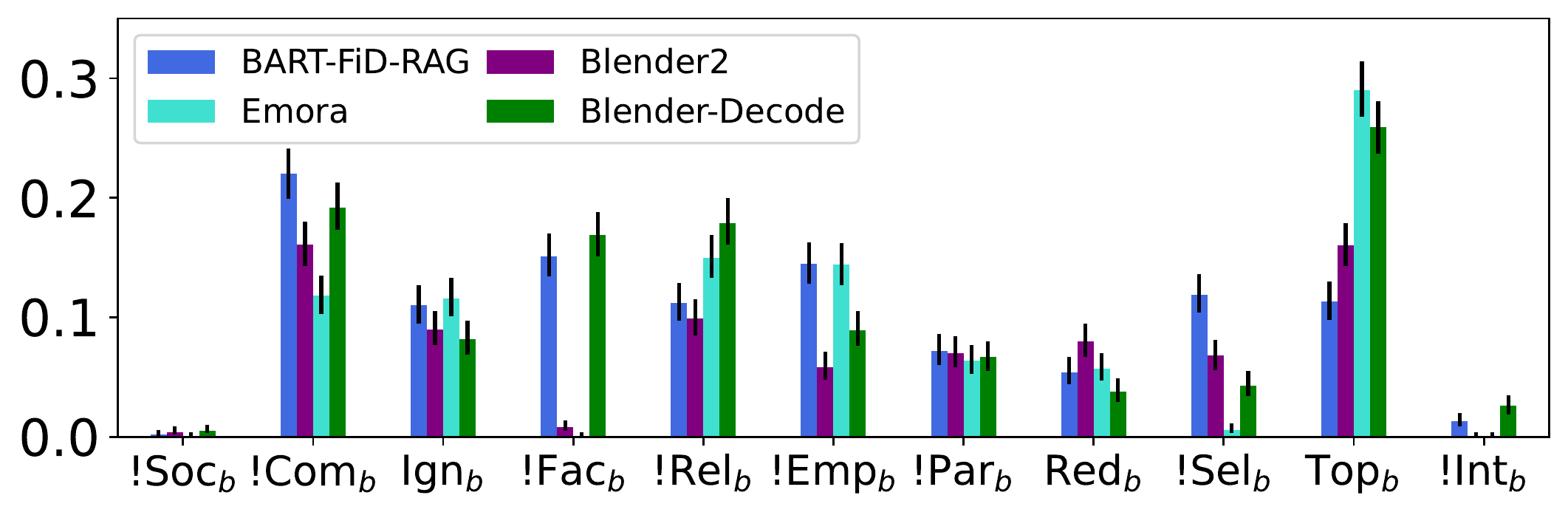}
    \vspace{-2ex}
    \caption{Proportions of turns expressing undesirable behaviors, with 95\% Wilson score confidence intervals.}
    \label{fig:undesirable_behaviors}
    \vspace{-2ex}
\end{figure*}

\begin{figure}[H]
    \centering
    \includegraphics[width=\columnwidth]{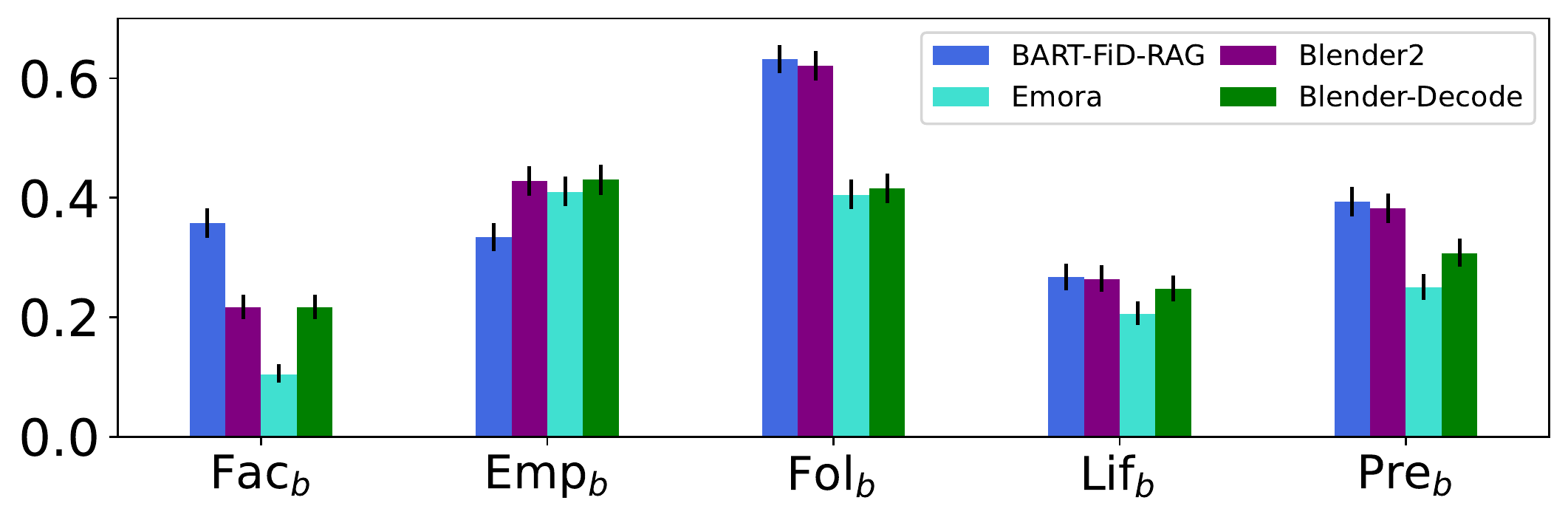}
    \vspace{-4.0ex}
    \caption{Proportions of turns expressing desirable behaviors, with 95\% Wilson score confidence intervals.}
    \label{fig:desirable_behaviors}
    \vspace{-3ex}
\end{figure}

\begin{figure}[H]
    \centering
    \includegraphics[width=\columnwidth]{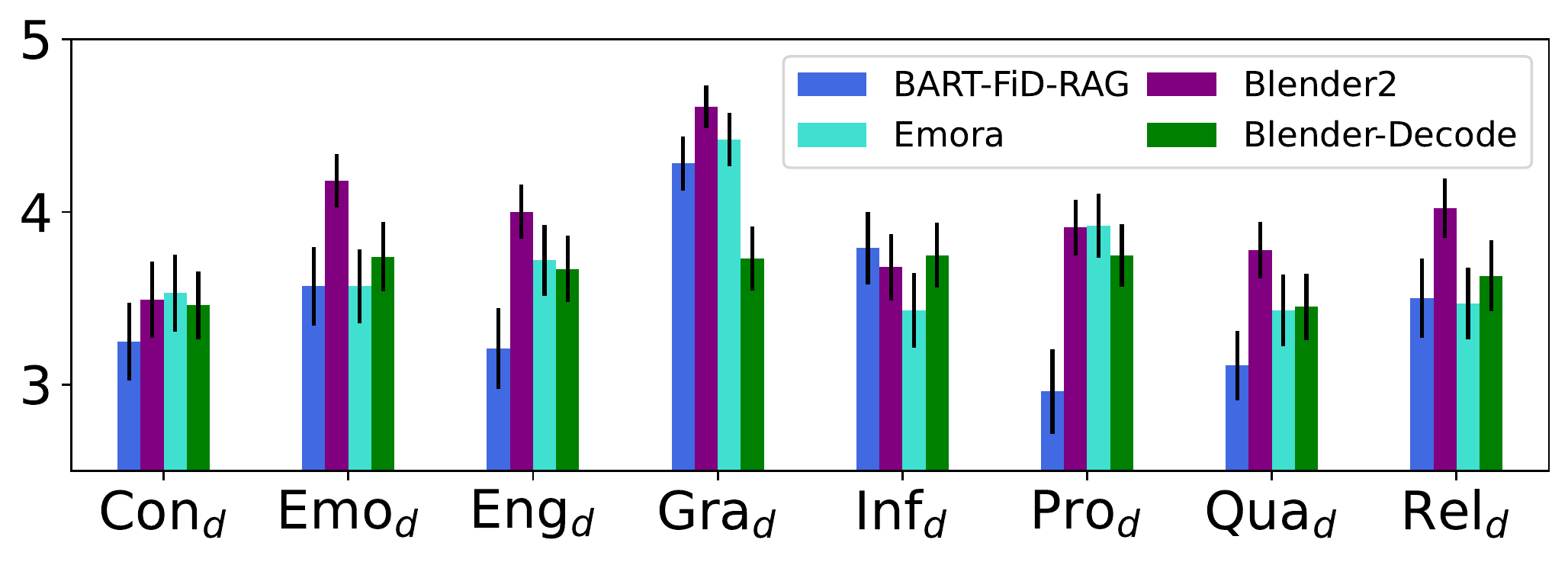}
    \vspace{-4.0ex}
    \caption{Average Dialogue Likert ratings of the conversations, with 95\% Student's t confidence intervals.}
    \label{fig:dialogue_likert}
    \vspace{-3ex}
\end{figure}

\begin{figure}[H]
    \centering
    \includegraphics[width=\columnwidth]{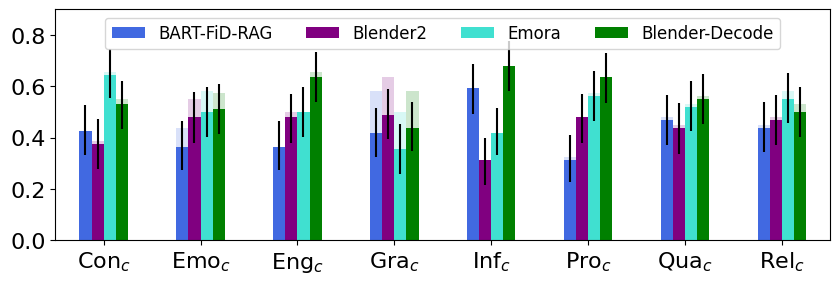}
    \vspace{-4.0ex}
    \caption{Win/tie rates of each bot vs all other bots, with 95\% Wilson score confidence intervals for win proportion. Transparent segments denote tie rates.}
    \label{fig:comparative}
    \vspace{-3ex}
\end{figure}

\begin{figure}[H]
    \centering
    \includegraphics[width=\columnwidth]{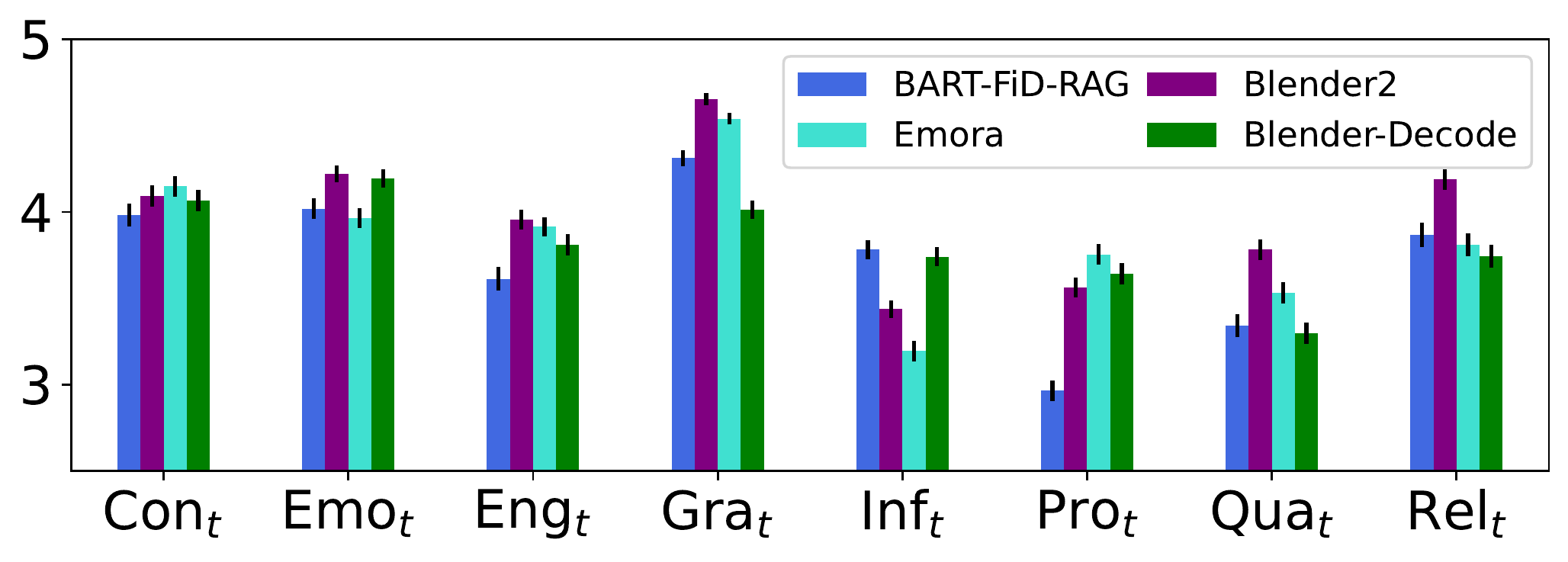}
    \vspace{-4.0ex}
    \caption{Average Turn Likert ratings of the conversations, with 95\% Student's t confidence intervals.}
    \label{fig:turn_likert}
    \vspace{-1ex}
\end{figure}

To evaluate the strengths and weaknesses of our 4 selected chatbots, we present results for the 400 collected conversations across all ABC-Eval metrics (Figure \ref{fig:undesirable_behaviors} and Figure \ref{fig:desirable_behaviors}), Likert Dialogue metrics (Figure \ref{fig:dialogue_likert}), Likert Turn metrics (Figure \ref{fig:turn_likert}), and Comparative metrics (Figure \ref{fig:comparative}). We focus our discussion on the final set of ABC-Eval metrics since they performed best in our metric analysis.

The results highlight the notable recent progress in human-computer chat, as the vast majority of chatbot turns are interpretable, relevant responses to the dialogue context. Less than 1\% of responses have interpretability issues, and Blender2 and BART-FiD-RAG each achieve a relevant response rate of nearly 90\%. Blender2 specifically is also able to incorporate factual knowledge into about 20\% of its responses while hallucinating factual information at a remarkably low rate, less than 1\%. Furthermore, the chatbots almost never produce responses with offensive language.\footnote{Note that our experiments are conducted with cooperative human interactors. Chatbots similar to those we test have been shown to reliably produce offensive language when responding to provocative inputs \cite{dinan2022safetykit}.} The chatbots also show a high rate of emotional understanding, with 40\% of their responses containing emotionally-appropriate reactions to the user. 

Despite these strengths, our results also show several clear directions for improvement. Commonsense violations are present in about 15-20\% of the bots' responses. Consistency issues are prevalent across all bots: self-contradictions, partner contradictions, and redundancies appear in about 5\% of the bots' responses overall. Also, all chatbots have a substantial rate of violating natural dialogue structure: about 10\% of responses are judged as ignoring the user, and depending on the chatbot, around 10-20\% of responses include irrelevant contributions to the dialogue. Additionally, 5-15\% of the chatbots' responses show a lack of empathy or other emotional misunderstandings. The reality of these observed rates of problematic behaviors is that, in most 30-turn conversations with these chatbots, a human interactor is likely to experience several issues that impact conversation quality.

\vspace{-1.5ex}

\vspace{-1.5ex}

\vspace{-1.5ex}

\section{Conclusion}
\label{sec:discussion}

As illustrated here, dialogue quality is a complex construct with many dimensions. 
Depending on the approach, dialogue systems can have markedly different weaknesses among these quality dimensions. 
Our research highlights several outstanding challenges, especially regarding the relevance, consistency, common-sensibility, and emotional understanding of chat model responses. 
Our analyses not only demonstrate that these four dimensions have a high impact on conversation quality, but also that current chatbots have substantial response error rates in these areas.

To efficiently address the challenges facing open-domain dialogue models, we need a reliable, dimensional evaluation method; however, our results show that popular evaluations such as dialogue-level Likert and comparative methods may not be suitable. 
The presented ABC-Eval serves as a promising alternative in this direction. Although the popular dialogue-level likert evaluation method may be the most cost-effective and robust method for measuring overall dialogue quality, we recommend that researchers additionally use the final set of ABC-Eval metrics, or a subset relevant to their scientific goals, to evaluate specific strengths and weaknesses of new chat models. Overall, we hope future work can use insights from our study to make better-informed decisions about which evaluation method to use, and to tackle the challenges facing current chatbots.

\section{Limitations}








There are several characteristics of the presented analyses that limit the scope of conclusions that can be drawn. We discuss how each of these limitations affect the takeaways of our results below. 

\paragraph{Number of Chatbots}

The generalizability of our metric analysis results (Section \ref{sec:metric-analysis}) is constrained by the fact that we were only able to include conversations from 4 chatbots in our analyses. We did our best to choose chatbots representative of the field and seem to have selected a fairly diverse group of models (Section \ref{sec:bot_evaluation_results}). However, it is possible that not all results we found in our metric analyses will generalize when evaluating other chat models. One possible example is the number of partner contradictions we observed among our 4 chatbots (Figure \ref{fig:undesirable_behaviors}), which may be similar by coincidence. If other chatbot models indeed differ more substantially in partner contradiction rates, our sensitivity metric analysis may have underestimated the sensitivity of our partner contradiction metric (Section \ref{sec:sensitivity}). In general, including a larger number of chatbots in a metric analysis will improve the chance that its results will apply to new chatbot models. Future work that performs metric analyses like those we presented, but with different chatbots than the 4 selected in this work, would aid further analysis of our results' generalizability.

\paragraph{Use of Surgers as Evaluators}

We perform our analyses using only a single evaluator group (Surgers). This choice of evaluator group does not harm the replicability of our methods, as other researchers have access to use of SurgeHQ or similar third-party annotation companies. However, several other evaluator groups are more popularly used for chat model evaluation, such as university students and Amazon Mechanical Turkers (MTurkers). We attempted to carry out our study with three evaluator groups (see Appendix \ref{sec:worker_analysis} for details), but were unable to proceed with student and MTurker evaluator groups due to time constraints. Consequently, it is unclear to what extent our metric analysis results will generalize to other choices of evaluator. 

\paragraph{Number of Collected Conversations}

As with any study involving a sampling procedure, resource constraints limit the number of collected samples, which in turn limits the statistical power of the study's analyses. Our study included 400 conversations, which provided more than adequate statistical power for most of our analyses. For example, our investigation of each metric's predictive validity (Section \ref{sec:importance}) relied on a simple linear regression analyses. At a significance level of $\alpha$=0.05, our 400 conversation samples would yield a statistical power of 1-$\beta$=0.80 to detect effect sizes of $f^2$=0.14$^2$ by F-test for each metric's regression. However, our analyses with the weakest statistical power are our dialogue-level analyses that compare bots with only 100 samples per bot. At 100 samples per bot, and assuming a standard deviation of 1.0 Likert points,\footnote{\citet{smith:22} reports standard deviations of Likert metrics between 0.8 and 1.3} a two-tailed t-test of mean Dialogue Likert rating would have a statistical power of 1-$\beta$=0.80 to detect differences of an effect size of Cohen's $d$=0.40. This is still a reasonable amount of statistical power, but leaves room for our study to produce inconclusive results when the true differences between chatbots are small.

\section{Ethics Statement}

The presented work aims towards improving the scientific methodology of chat model evaluation. To this end, we present a battery of analyses comparing several aspects of metric validity for four different evaluation methods (Section \ref{sec:metric-analysis}). Our results allow other researchers in the field to make better-informed decisions regarding appropriate evaluation methodology in human-computer chat. To ensure replicability of our methods we publicly release the annotation software and chatbot implementations used to collect our conversation and evaluation data. Additionally, we provide full transparency in our analyses by releasing the code for all our presented analyses. Finally, to aid future research efforts in human-computer chat modelling and evaluation, we release an anonymized version of our conversation and evaluation data. 

One ethical consideration involved in our work involved managing human workers in our data collection processes. All worker participation in our study was voluntary and involved zero subjective screening processes, with a complete description of worker tasks, workload, and timeframe provided before work was assigned. Workers could opt out of our study at any time for any reason. As compensation for work completed, we targeted a compensation rate of \$10/hour for student\footnote{Students' compensation is given as an Amazon Gift Card for convenience; students are informed of this prior to any work being completed} and Amazon Mechanical Turk workers, and a rate of \$20/hour for Surgers. We compensated on a per-task-completed basis to ensure timely completion of work, but verified that target hourly rates were reasonably approximated throughout the course of the study by measuring workers' median task completion times (see Appendix \ref{sec:collection_cost} for details). These measures ensured that all human work in our study was fair, transparent, and mutually-beneficial.

Other ethical considerations arise in our study's conversation collection. Unlike the collection of evaluation or annotation data, collecting interactive conversation data from human-computer interaction poses a small but meaningful risk that sensitive, damaging, or personally identifying information could get collected. We mitigated this risk in three ways. First, students were notified in multiple email communications and before each conversation that their conversations with our chatbots would be publicly released. Included in these notices was the instruction to refrain from releasing any personally identifiable or damaging information. Our instructions suggest that students fabricate personal information at any time during the conversations if it would make them feel more comfortable.
Second, we hand-checked all 400 conversations to ensure the non-presence of any sensitive information. Third, we anonymize all data before public release. Our study's collection and analysis of conversation data did not investigate interactors as human subjects, and we did not seek institutional review board approval. 

Finally, there is a concern in our study about the potential of the chatbots to respond to student interactors with toxic, insensitive, or vulgar language. The data-driven nature of some of our evaluated chat models means the chatbots are prone to reflecting any biases, toxicity, and vulgarity present in the training data (see \citet{dinan2022safetykit} for a quantitative analysis). A high rate of antisocial behaviors among our evaluated models could potentially make human interactors' experience talking with the bots quite uncomfortable, and would poorly reflect on the research field's potential for social good. To mitigate this risk, the authors extensively hand-tested all evaluated chat models, as well as conducting a pilot evaluation among the authors' lab group. As confirmed further in our results (Section \ref{sec:bot_evaluation_results}), our chatbots exhibited negligible rates of antisocial behavior.

\section{Acknowledgements}

We gratefully acknowledge the support of the Amazon Alexa AI grant. Any opinions, findings, and conclusions or recommendations expressed in this material are those of the authors and do not necessarily reflect the views of Amazon.
In addition, thank you to Bradley Webb, Scott Heiner, and the rest of the SurgeHQ team for their guidance in running our annotation projects on their platform.
We are also grateful to our colleagues at Emory for their participation in piloting the bots and refining the annotation interfaces. 
Lastly, a thank you to our reviewers for their helpful feedback.

\bibliography{anthology,custom}
\bibliographystyle{acl_natbib}

\clearpage
\newpage

\appendix

\section{Chatbot Selection Details}
\label{sec:chatbot_selection}


This appendix discusses the details of our literature review for each of the four chosen research themes.

\paragraph{General}
Our work focuses primarily on open-domain chat. 
Large-scale language modeling using dialogue-like pretraining data can produce surprisingly human-like conversation on virtually any popular conversation topic \cite{roller:21}.
Of these approaches, we chose Blender2 \cite{weston:21}, which reportedly outperformed its previous iteration Blender \cite{roller:21} who had surpassed DialoGPT \cite{zhang:20} and Meena \cite{adiwardana:20}. 

There had been also several chatbots produced by the Amazon Alexa Prize Socialbot Grand Challenge \cite{ram:18} focusing on general, open-domain chat, most of which incorporate rule-based methods to ensure interesting and consistent responses \cite{khatri:18}. 
Since these chatbots performed well in practice but lack comparison to SOTA data-driven models, we selected the bot with the all-time highest final score, Emora\footnote{\href{https://github.com/emora-chat/emora_ap3_parlai}{https://github.com/emora-chat/emora\_ap3\_parlai}} \cite{finch_emora:20}, as one of our candidates.

\paragraph{Knowledge}
Grounding chat with supplementary knowledge resources is a common way to improve engagingness and control the topic of conversation \cite{li_incremental:19, ye:20}. 
ColV \cite{zhan:21} achieved SOTA performance in knowledge-grounded dialogue response generation on the popular WoW dataset \cite{dinan:19}; however, no implementation was publicly available.
DukeNet \cite{meng:20} and PIPM \cite{chen:20} report next-best performance in this task. 
DukeNet's implementation was available while PIPM's was not, therefore we selected DukeNet as a candidate. 

BART-FiD-RAG also reported compelling performance for knowledge-grounded chat \cite{shuster:21}, but did not compare to other SOTA models we identified. 
Since BART-FiD-RAG’s inclusion in ParlAI provided easy replication, we included it in our bot pilot.

\paragraph{Consistency}
Improving consistency of chatbot responses is noted as a challenge and addressed in several works \cite{welleck:19, nie:21, li:21}. 
DECODE \cite{nie:21} reported SOTA performance for general inconsistency avoidance, improving upon an approach that used unlikelihood training with dialogue natural language inference data \cite{li:20}. 
Note that there were several works focusing specifically on persona consistency \cite{song_generate:20, kim:20, song:21}, which we did not consider due to their narrower contradiction scope.

\paragraph{Empathy}
Several works demonstrated the importance of emotional understanding in chat \cite{partala:04, prendinger:05, kim:21, sabour:22}. 
To provide contrast with our knowledge-grounded candidates, we selected CEM \cite{sabour:22}, which reported SOTA results in empathetic response generation. 
Many related works investigated controllable emotional response generation \cite{song:19, zhong:21}, but we did not consider models requiring an emotion label as input.


\section{Chatbot Implementation Details}
\label{sec:chatbot_implementations}

For each selected candidate model, a brief overview of the implementation details required to use them as interactive models in this work is below:

\paragraph{Emora}
We implement a ParlAI agent using the interactive chatting mode provided for the Emora system \cite{finch_emora:20}.

\paragraph{BART-FiD-RAG}
An interactive chatting mode for BART-FiDRAG is provided through ParlAI. We use the default arguments, except for \texttt{compressed-indexer-nprobe} and \texttt{beam-context-block-ngram} which we set to 128 and 3, respectively, following internal testing by the authors before the pilot.

\paragraph{Blender2}
An interactive chatting mode for Blender2\footnote{\href{https://parl.ai/projects/blenderbot2/}{https://parl.ai/projects/blenderbot2/}} is provided through ParlAI; however, the internet-search augmented functionality as described in \citet{Komeili:22} is not provided. Although there is an available alternative internet search server\footnote{\href{https://github.com/JulesGM/ParlAI_SearchEngine}{https://github.com/JulesGM/ParlAI\_SearchEngine}} that is compatible with Blender2, it increased the response latency of Blender2 such that it no longer satisfied our response latency requirement. As a result, we use the 3B parameter version of Blender2 that does not include the internet search functionality.

\paragraph{Blender-Decode}
We implement a custom agent in the Parlai framework following the best-performing setup of \citet{nie:21} of finetuning a consistency classification model and using it to rerank generated responses from Blender \cite{roller:21} such that the least contradictory response is selected.

\paragraph{DukeNet}
The published code for DukeNet is missing an interactive chatting mode, although a trained model on the multi-topic Wizard of Wikipedia (WoW) \cite{dinan:19} is provided. We modified the published code to support an interactive chatting mode. However, one major difference is that the relevant knowledge documents need to be retrieved in interactive chatting mode, whereas they are predetermined in the training and evaluation environments. To remedy this, we use a trained Dense Passage Retrieval model \cite{karpukhin:20} provided through Sentence-Transformers\footnote{\href{https://www.sbert.net/docs/pretrained-models/dpr.html}{https://www.sbert.net/docs/pretrained-models/dpr.html}} to select knowledge at each dialogue turn. We implement a ParlAI agent using the available DukeNet model and our custom interactive chatting mode.

\paragraph{CEM}
The published code for CEM is missing both an interactive chatting mode and a trained model. We trained a CEM model using the default arguments from the repository and modified the published code to support an interactive chatting mode. We implement a ParlAI agent using this trained model and our custom interactive chatting mode. 

\begin{table}[ht!]
\centering
    \begin{tabularx}{\columnwidth}{|l|X|}
    \hline
    \bf User          & Hi! \\
    \bf Bot          & What is happening? \\
    \bf User          & I am just getting my day started. Finished breakfast and booted up my computer to begin working. What about you? \\
    \bf Bot & That is great! I wish you the best of luck! \\
    \bf User & Yeah, I think it will be a good day and I am excited for it. What are you doing today? \\
    \bf Bot & I am sure you will be fine! \\
    \bf User & I'm not too worried about it. \\
    \bf Bot & I am so happy for you! \\
    \bf User & So, what do you do for work? \\
    \bf Bot & I am sure you will do great! \\
    \hline
    \end{tabularx}%
  \caption{Chatbot pilot conversation: CEM}
  \label{tab:cem_convo}
\end{table}

\begin{table}[ht!]
  \centering
    \begin{tabularx}{\columnwidth}{|l|X|}
    \hline
    \bf User          & Hi! \\
    \bf Bot          & Hello, how are you? \\
    \bf User          & I am pretty good what about you \\
    \bf Bot & I'm doing well, thanks for asking. How's the weather in your neck of the woods? \\
    \bf User & pretty warm \\
    \bf Bot & That's good. Do you have any plans for the weekend? I'm thinking of going to the beach. \\
    \bf User & cool I have a gathering with my church friends \\
    \bf Bot & That sounds like a lot of fun. I haven't been to church in a long time. How long have you been going to church? \\
    \bf User & many years since I was very young \\
    \bf Bot & That's cool. I don't think I've ever been to a church service in my life. \\
    \bf User & what do you do for fun \\
    \bf Bot & I like to watch movies and play video games. What do you like to do in your free time? \\
    \hline
    \end{tabularx}%
  \caption{Chatbot pilot conversation: Blender2}
  \label{tab:blender2_convo}
  \vspace{-3ex}
\end{table}

\section{Bot Pilot Examples}
\label{sec:bot_pilot_examples}

Tables \ref{tab:cem_convo} and \ref{tab:blender2_convo} show two of our chatbot pilot (Section \ref{sec:chatbot_selection}) conversations, one from CEM \cite{sabour:22} and one from Blender2 \cite{weston:21}, that exemplify the difference between single-turn and multi-turn dialogue response generation models. The CEM model is trained to give an empathetic response to a dialogue context, and achieves good performance towards this goal. However, as shown in the example, this response policy does not translate well for multi-turn interaction with a human. By contrast, Blender2 is trained and evaluated specifically to achieve multi-turn dialogue.

\section{Pilots and Development}
\label{sec:label_pilots}
The final ABC-Eval label set, annotation procedure, and software application are created using an iterative process of development and piloting. 
14 students are invited to serve as evaluators for piloting the evaluation. 
To avoid overfitting the evaluation design, our pilots evaluated conversations collected between Blender \cite{roller:21} and one of the authors, and a new set of conversations was used for each pilot round. 
We ran 4 pilot rounds, making revisions after manually reviewing each round's annotation.

\begin{table}[htbp]
\resizebox{\columnwidth}{!}{%
\begin{tabular}{l|c|c|l|c}
\textbf{} & \textbf{Dialogues} & \textbf{Annotators} & \textbf{Type} & \textbf{$\alpha$} \\ \hline
\multicolumn{1}{c|}{\multirow{2}{*}{Pilot 1}} & 4 & 11 & Lab (ALL) & 0.18 \\ \cline{2-5} 
\multicolumn{1}{c|}{} & 2 & 4 & Lab (G) & 0.39 \\ \hline
\multirow{2}{*}{Pilot 2} & 6 & 5 & Lab (ALL) & 0.49 \\ \cline{2-5} 
 & 6  & 4 & Lab (G) & 0.50 \\ \hline
\multirow{2}{*}{Pilot 3} & 6 & 4 & Lab (U) & 0.43 \\ \cline{2-5} 
 & 6 & 4 (screened) & Lab (U) & 0.45 \\  
\end{tabular}%
}
\caption{The distribution of dialogues and annotators for each annotation pilot. Pilot 1 included full label annotation of each dialogue where dialogues were distributed among groups of annotators. Pilots 2 and 3 had a subset of labels annotated for each dialogue, but all annotators annotated each dialogue.  \texttt{$\alpha$}: Krippendorf's alpha, \texttt{U}: Undergraduate annotators, \texttt{G}: Graduate level annotators, \texttt{ALL}: Undergraduate and graduate annotators.}
\label{tab:pilot_stats}
\end{table}

\noindent Table \ref{tab:pilot_stats} presents a summary of the major changes made in each pilot round and IAA metrics. It is important to note that each annotator performed all of the annotation tasks in one sitting in sequence for each pilot. These piloting rounds are not necessarily directly comparable to one another when taken as a whole, since the annotator groups and dialogues to be annotated varied between each round. Instead, we will discuss below the major takeaways afforded by different splits of the pilots that informed the final design of ABC-Eval. 

\paragraph{Subtask Formulation} 
The decision to format ABC-Eval into several small annotation subtasks, each with a tailored subset of the behavior labels, was made from the results of Pilot 1. In Pilot 1, we divided the initial set of annotation labels into 3 annotation subtasks each with 4-9 labels: errors, information usages (commonsense, world knowledge, etc.), and utterance types (request, presentation, etc.). Each annotator performed the annotation tasks in one sitting in sequence. The overall interannotator agreement was quite low ($\alpha=0.18$), which was concerning. Based on ad-hoc feedback from the pilot annotators, the consensus was that each subtask demanded an unreasonable cognitive load on annotators due to the large number of labels to keep track of. 

For Pilot 2 we increased the number of annotation tasks such that each covered a small and related scope of behavior labels, with 1-4 labels per task. Table \ref{tab:pilot_stats} shows the boost to interannotator agreement between Pilots 1 and 2. However, this agreement increase could have resulted from an increase in the quality of the annotators (as Pilot 2 was composed primarily of annotators with a graduate-level education whereas Pilot 1 was more evenly split between annotators with an undergraduate-level education and graduate-level education). To remove this confound, we calculated the agreement in Pilots 1 and 2 when only considering graduate-level annotators. Although it was less dramatic, there remained an increase in agreement from 0.39 to 0.50, which encouraged the decision to maintain the smaller annotation subtasks. Dividing the annotation into tailored subtasks seemed to reduce the cognitive load on annotators, thus allowing them to perform more accurate annotations per task.

\paragraph{Training and Screening}

Manual analysis of the pilot annotations from Pilots 1 and 2 revealed some recurring annotation mistakes, arising from misunderstandings of the guidelines for the tasks. In an attempt to correct such misunderstandings, a training procedure was introduced for each task.

Each round of training consists of 1 curated conversation with ground-truth labels and explanations that are shown as feedback to the annotator after they complete the training round. We used the results of Pilots 1 and 2 in order to develop these curated conversations as follows:

\begin{enumerate}
    \item \textbf{Label Specifications:} We constructed a label specification that consisted of a comprehensive enumeration of positive and negative cases of the label with the goal of defining a decision boundary that the annotators should strive towards. We especially focused on the utterances for which several of the annotators failed to produce labels that matched the ground truth annotations we had defined for each of the Pilots.
    \item \textbf{Training Conversation Selection:} We selected 3 conversations between Blenderbot and a human (from a collection within our lab) for each label to be used as training conversations for it. This selection was manually done by ranking the conversations on their coverage of the label specification.
    \item \textbf{Training Conversation Modification:} We heavily revised the selected conversations by hand to ensure that all of the cases identified in the specification were adequately represented, most often by inserting new utterances that corresponded to any underrepresented cases.
\end{enumerate}

\begin{figure*}[!ht]
\centering
\includegraphics[width=0.75\textwidth]{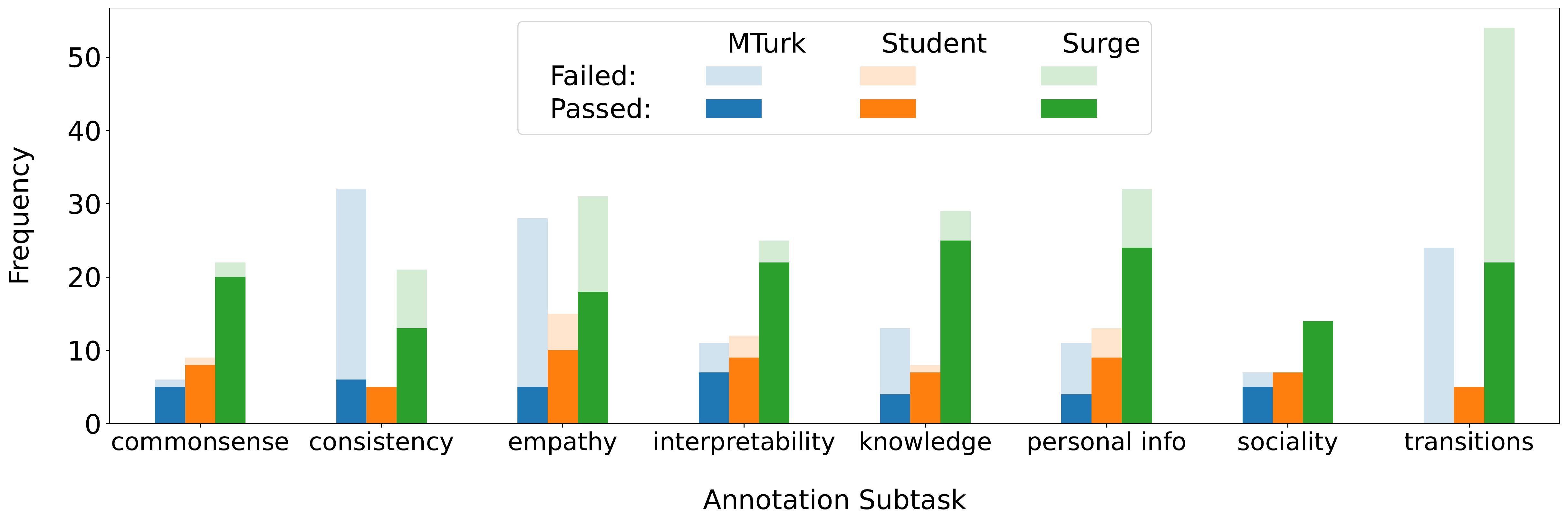}
\caption{The training pass rate of each annotation workforce for the behavior evaluation tasks.}
\label{fig:training_rates}
\end{figure*}

\noindent To evaluate the utility of this training process, a third pilot was conducted using 4 undergraduates. We observed a general upwards trend in annotation performance between the training rounds for the annotators, suggesting that the training was aiding in the annotation accuracy for the annotators. The final agreements were 0.43 and 0.45 between all annotators and annotators who passed the training, respectively, on the annotated conversations. 

Due to the small nature of this pilot, we are unable to conclude whether this difference is meaningful. However, ad-hoc feedback from the annotators suggested that the training rounds were useful towards their understanding of the tasks, although the amount of training did increase the overall workload of participation. Accordingly, the decision was made to treat each subtask independently, rather than require all subtasks to be completed for one dialogue in a single sitting for each annotator.

\paragraph{General Revisions}
Throughout each of these pilot rounds the annotation instructions, examples, and training rounds were updated based on manual review of the annotations in an attempt to correct any unclear or misleading information.

\section{Evaluator Training and Screening}
\label{sec:worker_analysis}

We attempted to use three different groups of evaluators for our full evaluation study:

\paragraph{Students}
Undergraduate students were recruited from the authors’ university via word-of-mouth and email advertisements sent to computer science, psychology, and quantitative methods departmental mailing lists.\footnote{Students were compensated with an Amazon gift card at the completion of the data collection.}

\paragraph{MTurkers}
Our 20 evaluation tasks were posted to the Amazon Mechanical Turk crowdsourcing platform.\footnote{\href{https://www.mturk.com/}{https://www.mturk.com/}} 

\paragraph{Surgers}
Our 20 evaluation tasks were posted on SurgeHQ’s annotation platform\footnote{\href{https://www.surgehq.ai/}{https://www.surgehq.ai/}} to be completed by dedicated workers with experience in NLP annotation. A group of 125 Surgers were qualified to participate in our tasks, chosen by a SurgeHQ employee on the basis of high annotation performance on past projects.

\vspace{1ex}
\noindent All three groups were compensated per task per annotated conversation, at an estimated rate of \$10/hr for Students and MTurkers, and \$20/hr for Surgers.

\noindent To check the viability of each worker group to produce evaluation data for our full study, we released a random 5 conversations out of our set of 400, to be fully evaluated by each worker group in each of our 8 ABC-Eval tasks. After a two week period, Surgers were the only worker group that were able to fully evaluate the 5 conversations in all 8 ABC-Eval tasks. This was due to an overall lack of participation from the Student group, and due to low training pass rates from the MTurk group (see Figure \ref{fig:training_rates} for quantitative outcomes). Although worker group differences in work rate and training performance might be explained by the difference in compensation structure, we decided to proceed with the Surgers group only for our full study to collect our evaluation data in a timely manner. 

\section{Collection Cost}
\label{sec:collection_cost}

Compensation rates are based on per-task completion times from an internal pilot run. The rates per task paid to Surgers are shown in Table \ref{tab:task_costs}. We also present the real and theoretical costs for collecting each method included in our evaluation data (Table \ref{tab:costs}). As expected, turn-level annotation tasks are an order of magnitude more expensive to collect than dialogue-level tasks. Notably, the final set of ABC-Eval labels (Table \ref{tab:behavior_labels}) are, on average, less expensive to collect than turn-level Likert labels.

\begin{table}[ht]
\centering
\resizebox{\columnwidth}{!}{%
\begin{tabular}{l|c||l|c}
\toprule
\multicolumn{1}{c|}{\textbf{Task}} & \multicolumn{1}{c||}{\textbf{Payment}} & \multicolumn{1}{c|}{\textbf{Task}} & \multicolumn{1}{c|}{\textbf{Payment}} \\
\midrule
Uninterpretable & \$0.63 & Antisocial & \$0.44 \\
\midrule
Preference Info & \multirow{2}{*}{\$0.70} & Empathetic & \multirow{2}{*}{\$1.15} \\ 
Life Info & & Lack of Empathy & \\
\midrule
Commonsense & \multirow{2}{*}{\$0.92} & Fact Usage & \multirow{2}{*}{\$1.96}\\ 
Contradiction & & Fact Contradiction &\\
\midrule
Self Contradiction & \multirow{4}{*}{\$0.87} & Ignore & \multirow{4}{*}{\$1.87}\\ 
Partner Contradiction &  & Irrelevant &\\
Redundant & & Follow-up &\\
&& Topic Switch &\\
\midrule
Dialogue Likert & \$0.60 & Turn Likert & \$0.70 \\
\midrule 
Comparative & \$1.43 & \\
\bottomrule
\end{tabular}
}
\caption{Payment per annotation task in USD. The payment for Turn Likert is per label whereas the indicated payment for Dialogue Likert and Comparative covers all labels, due to how the annotation tasks were constructed (Section \ref{sec:eval_study}). }
\label{tab:task_costs}
\end{table}

\begin{table}[htbp]
\centering
\resizebox{\columnwidth}{!}{%
\begin{tabular}{l|r|r|r|r}
\toprule
\multicolumn{1}{c|}{\bf Metric} & \multicolumn{1}{c|}{\textbf{TI}} & \multicolumn{1}{c|}{\textbf{TP}} & \multicolumn{1}{c|}{\textbf{EC}} & \multicolumn{1}{c}{\textbf{OC}} \\
\midrule
Dialogue Collection & 8.08 & 7.43 & 1077.14 & 333.33 \\ 
Dialogue Likert & 2.81 & 21.37 & 374.36 & 240.00  \\ 
Comparative & 4.35 & 13.81 & 289.68 & 286.67  \\ 
Turn Likert & 19.94 & 3.01 & 2658.40 & 2240.00  \\ 
ABC-Eval$_{all}$ & 25.60 & 2.34 & 3413.58 & 3422.67 \\ 
ABC-Eval$_{final}$ & 15.17 & 3.95 & 2022.98 & - \\
\bottomrule
\end{tabular}%
}
\caption{The data collection costs for each task in United States Dollars. \textbf{TI}me is the median completion time in minutes for one dialogue. \textbf{T}hrough\textbf{P}ut represents the number of completed dialogues per hour. \textbf{E}stimated \textbf{C}ost is calculated using median completion time, 400 dialogues, and \$20/hr rate. \textbf{O}ur \textbf{C}ost is the total amount paid in this work to collect a dataset of 400 conversations (single-annotated). }
\label{tab:costs}
\end{table}  


\section{Evaluation Interfaces}
\label{sec:interfaces}

\begin{figure*}[b]
    \centering
    \includegraphics[width=0.9\textwidth]{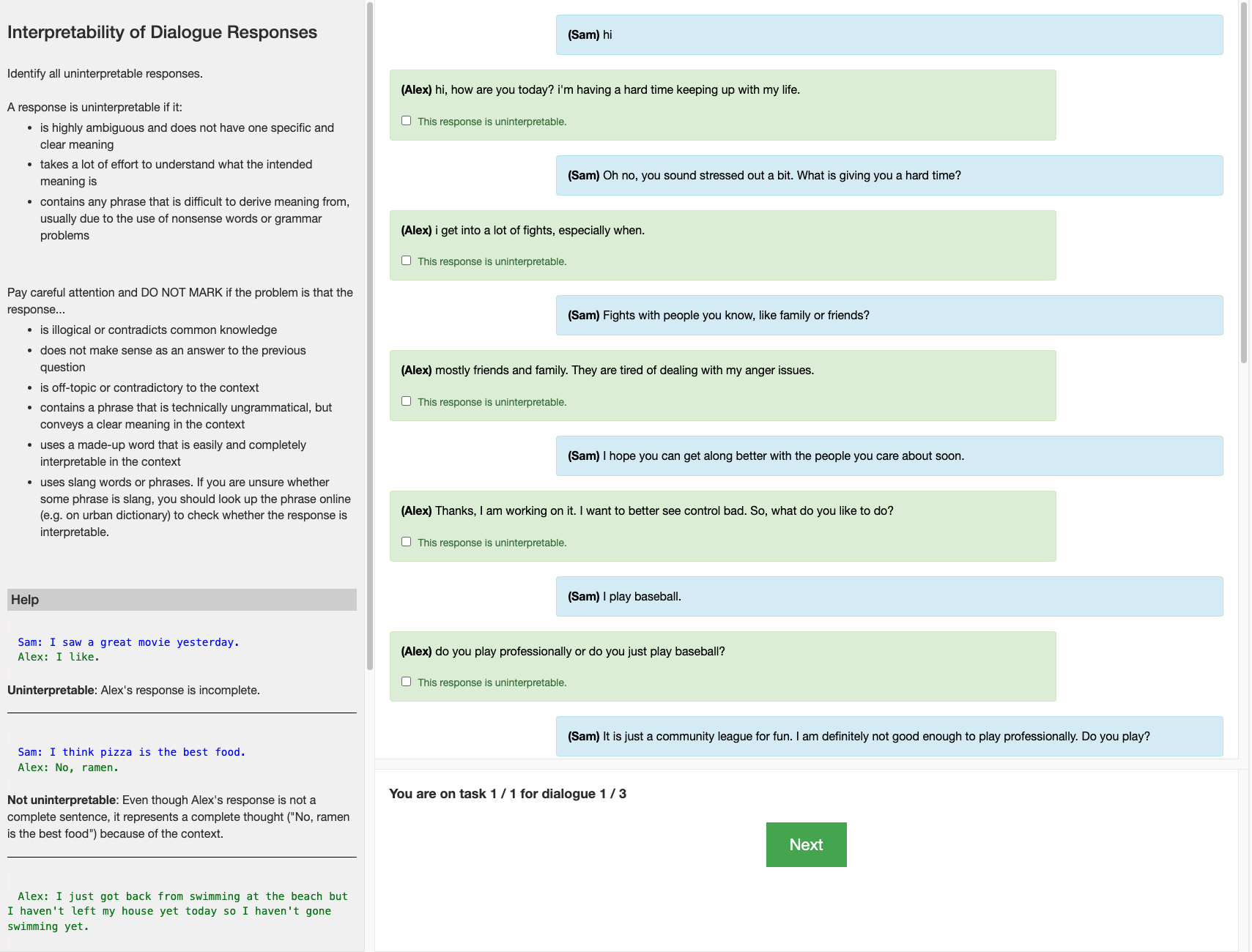}
    \caption{Interface for \texttt{uninterpretable}}
    \label{fig:interface_interpretability}
\end{figure*}

\begin{figure*}
    \centering
    \includegraphics[width=0.9\textwidth]{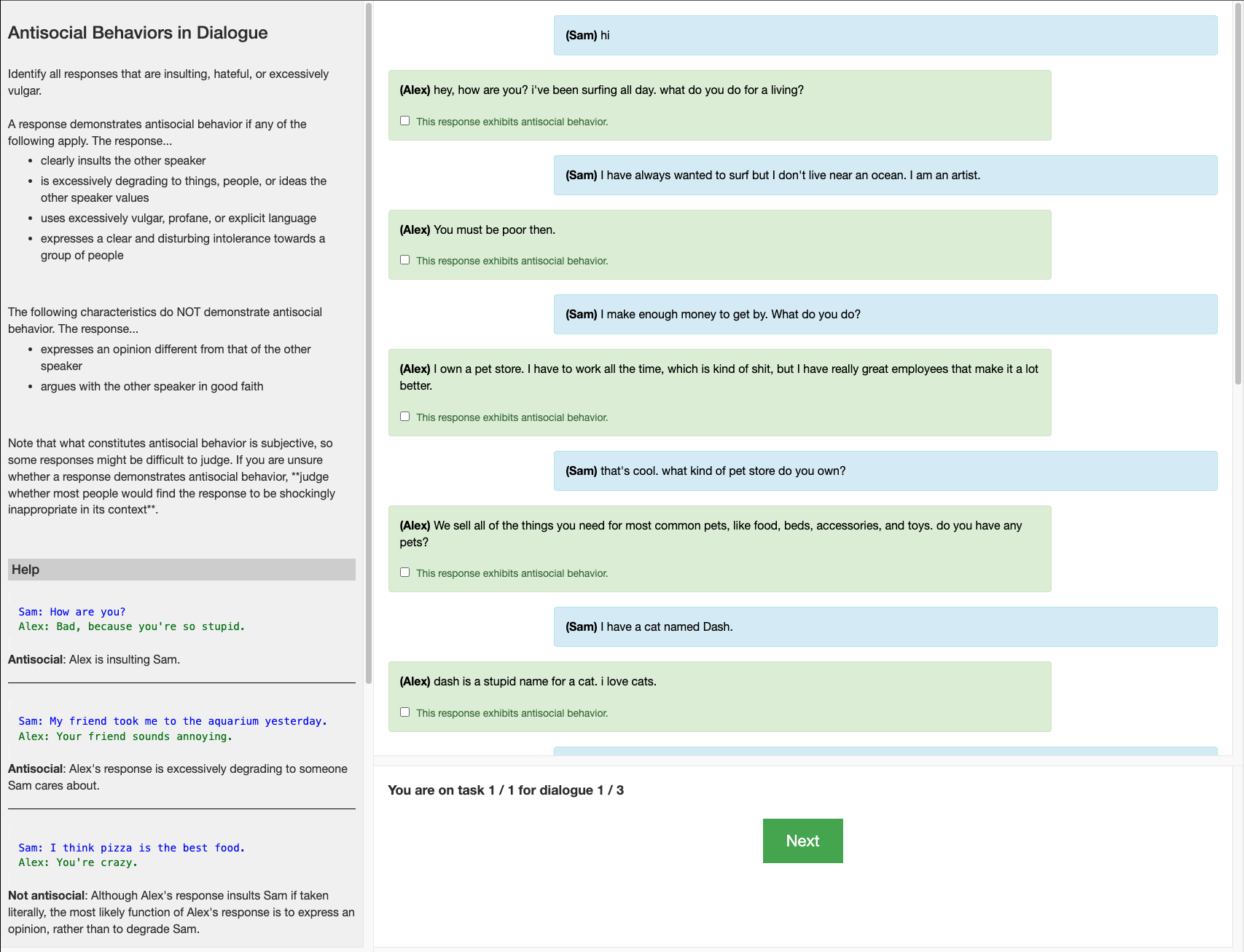}
    \caption{Interface for \texttt{antisocial}}
    \label{fig:interface_sociality}
\end{figure*}

\begin{figure*}
    \centering
    \includegraphics[width=0.9\textwidth]{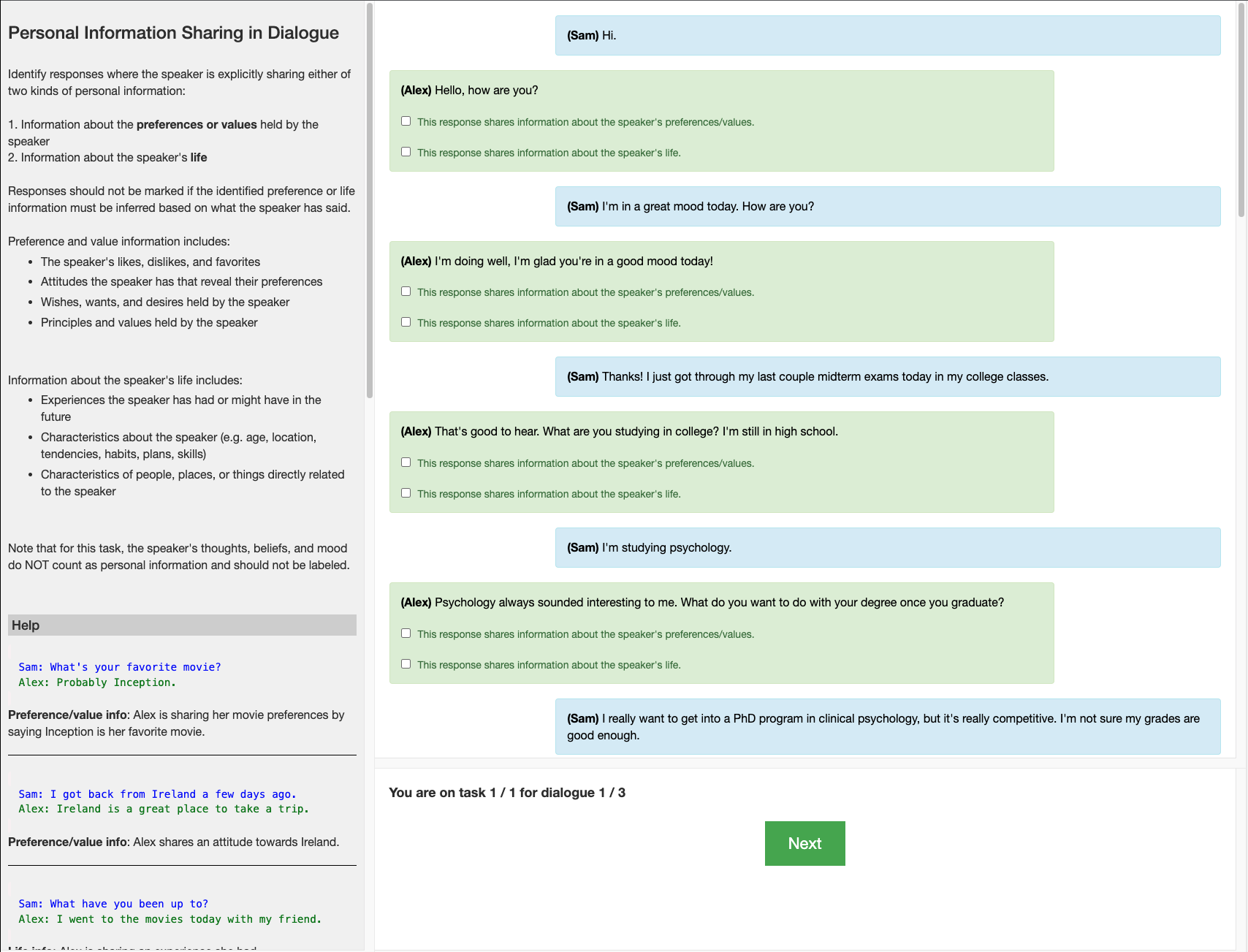}
    \caption{Interface for \texttt{preference info} and \texttt{life info}}
    \label{fig:interface_personalinfo}
\end{figure*}

\begin{figure*}
    \centering
    \includegraphics[width=0.9\textwidth]{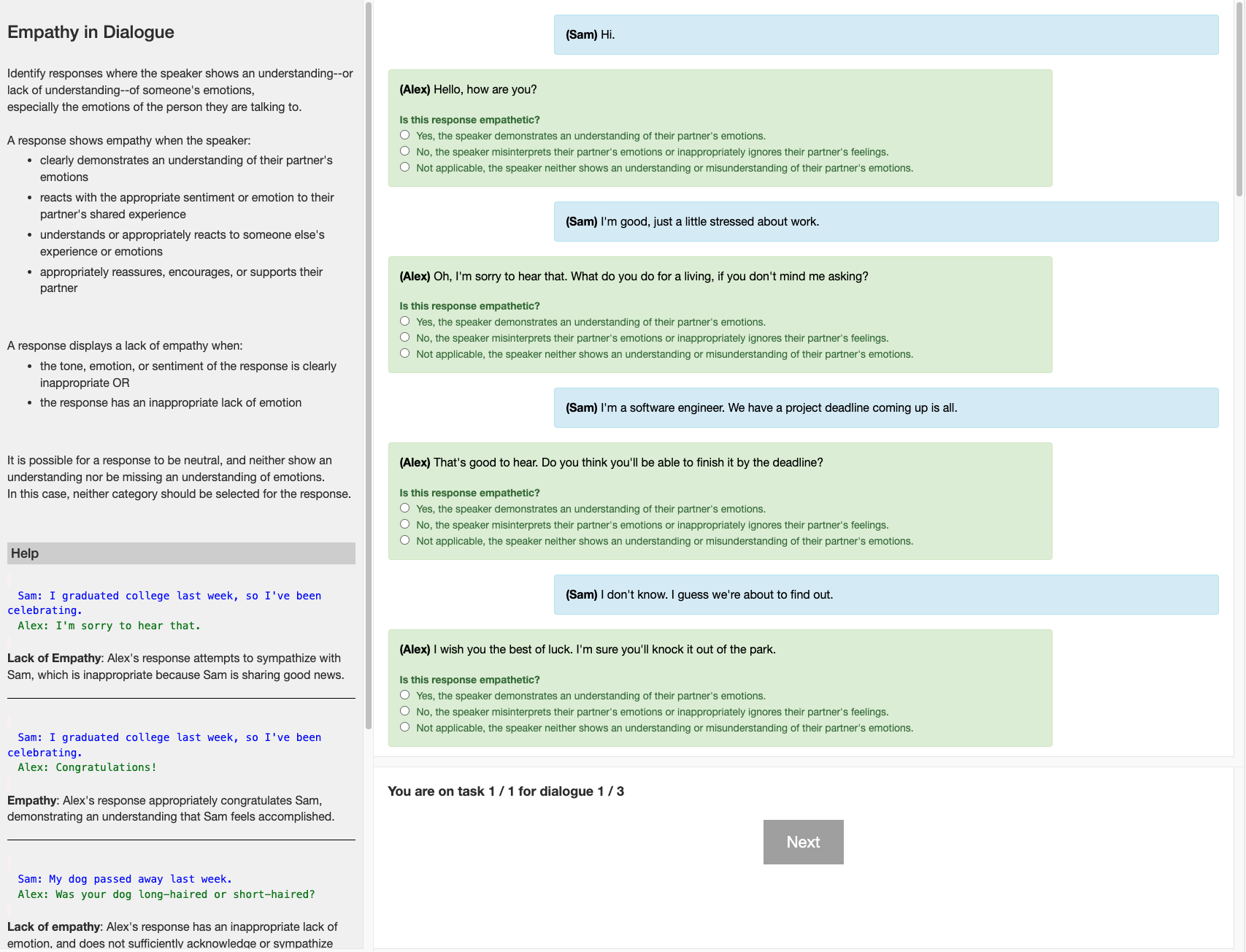}
    \caption{Interface for \texttt{empathetic} and \texttt{lack of empathy}}
    \label{fig:interface_empathy}
\end{figure*}

\begin{figure*}
    \centering
    \includegraphics[width=0.9\textwidth]{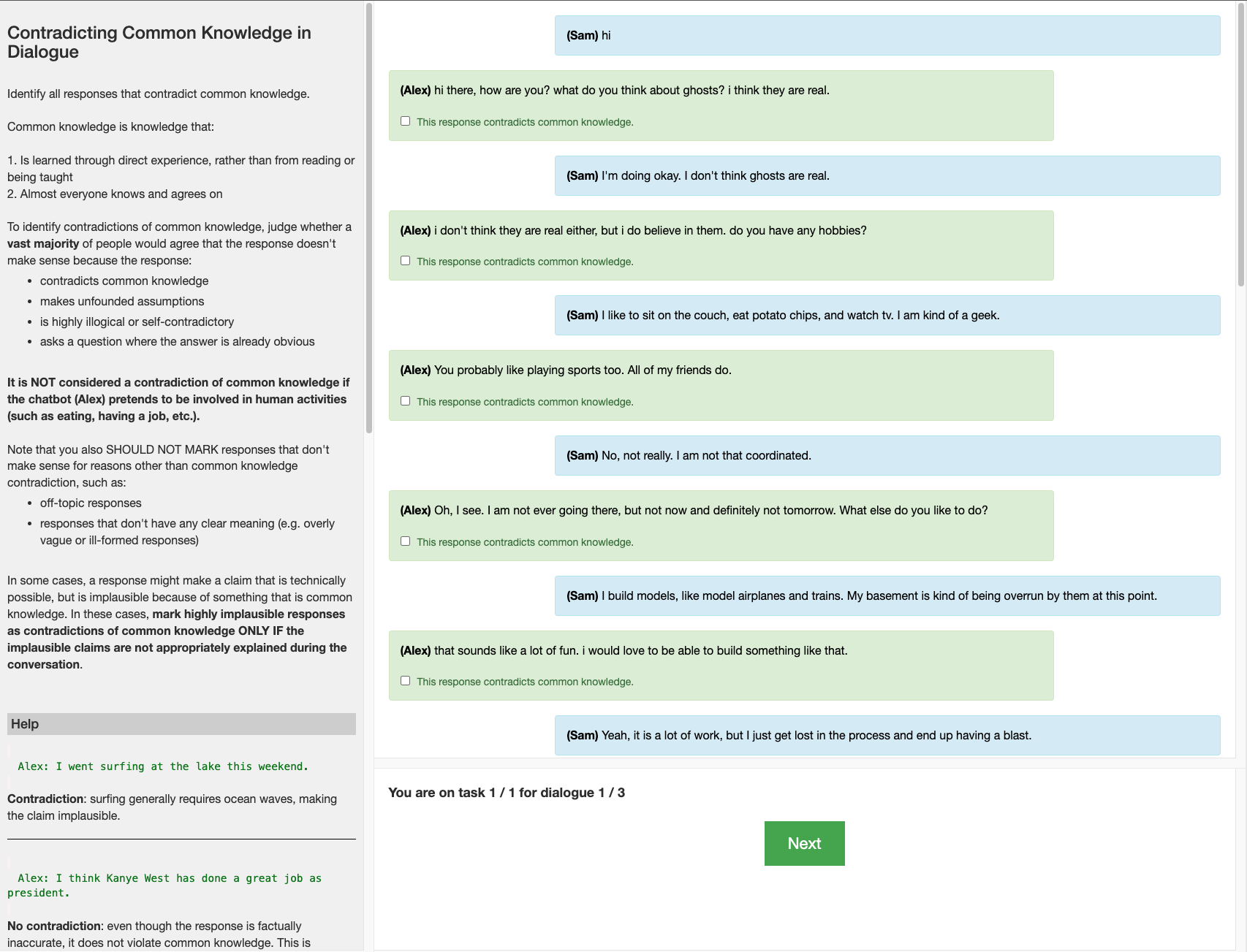}
    \caption{Interface for \texttt{commonsense contradiction}}
    \label{fig:interface_commonsense}
\end{figure*}

\begin{figure*}
    \centering
    \includegraphics[width=0.9\textwidth]{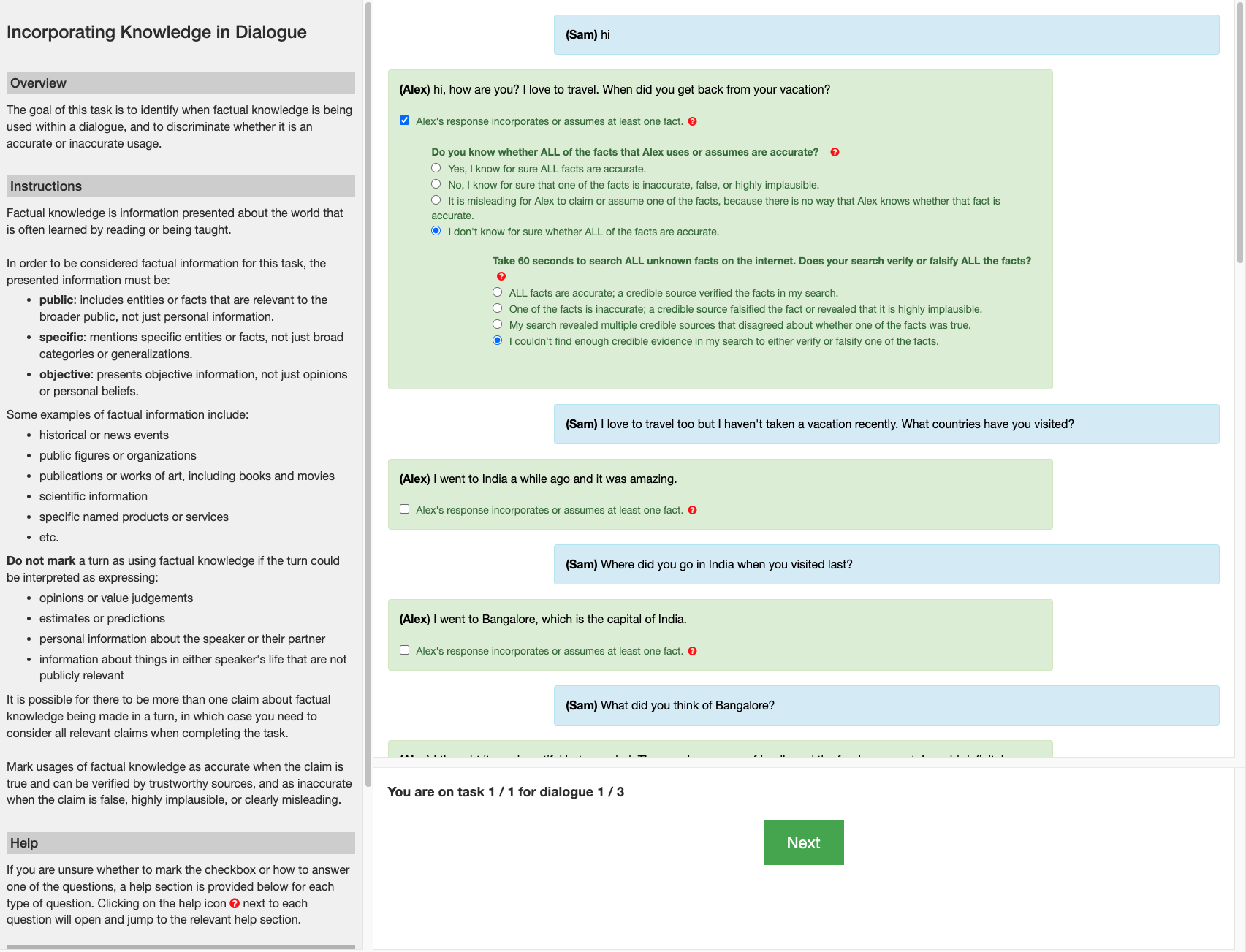}
    \caption{Interface for \texttt{fact usage} and \texttt{fact contradiction}}
    \label{fig:interface_knowledge}
\end{figure*}

\begin{figure*}
    \centering
    \includegraphics[width=0.9\textwidth]{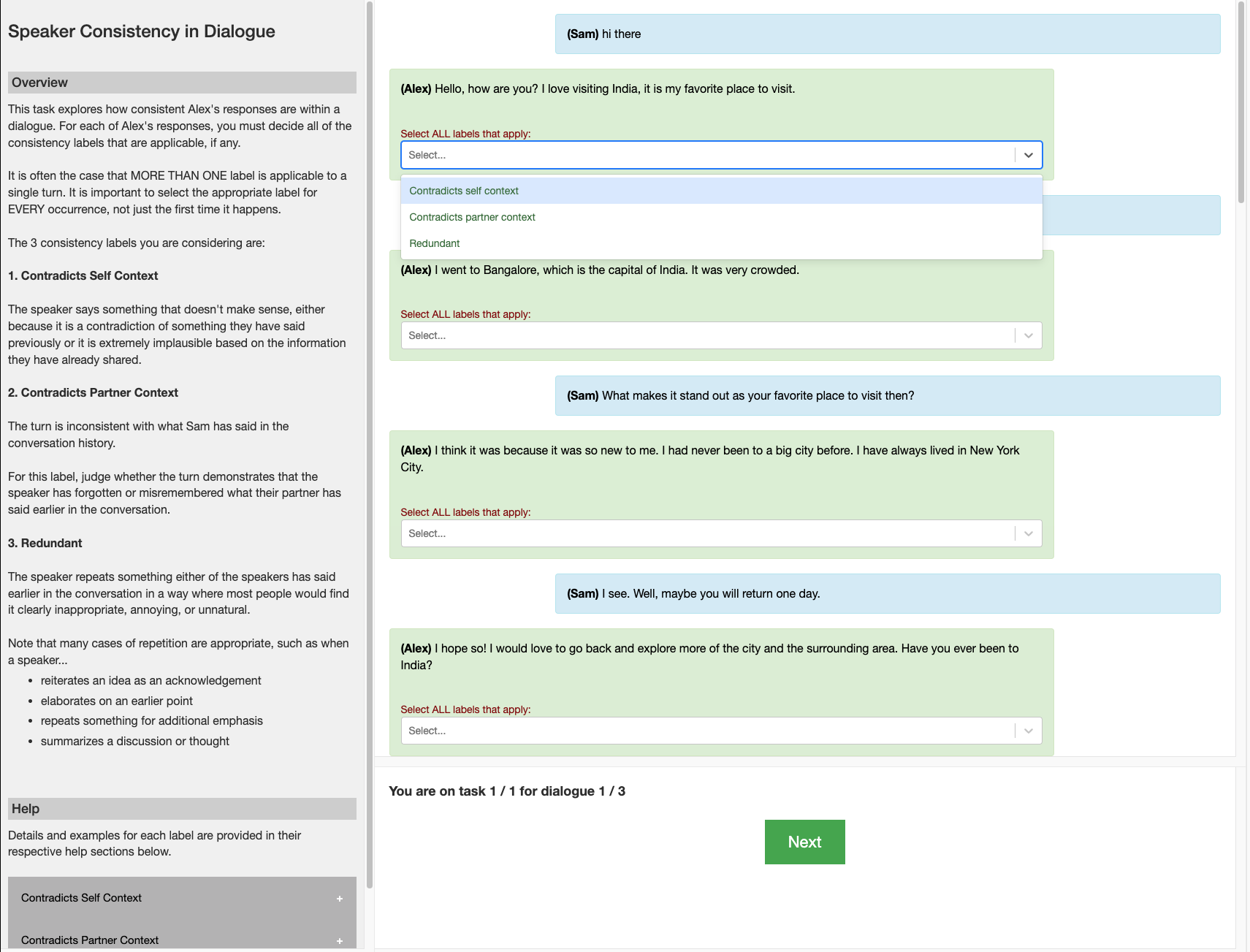}
    \caption{Interface for \texttt{self contradiction}, \texttt{partner contradiction}, and \texttt{redundant}}
    \label{fig:interface_consistency}
\end{figure*}

\begin{figure*}
    \centering
    \includegraphics[width=0.9\textwidth]{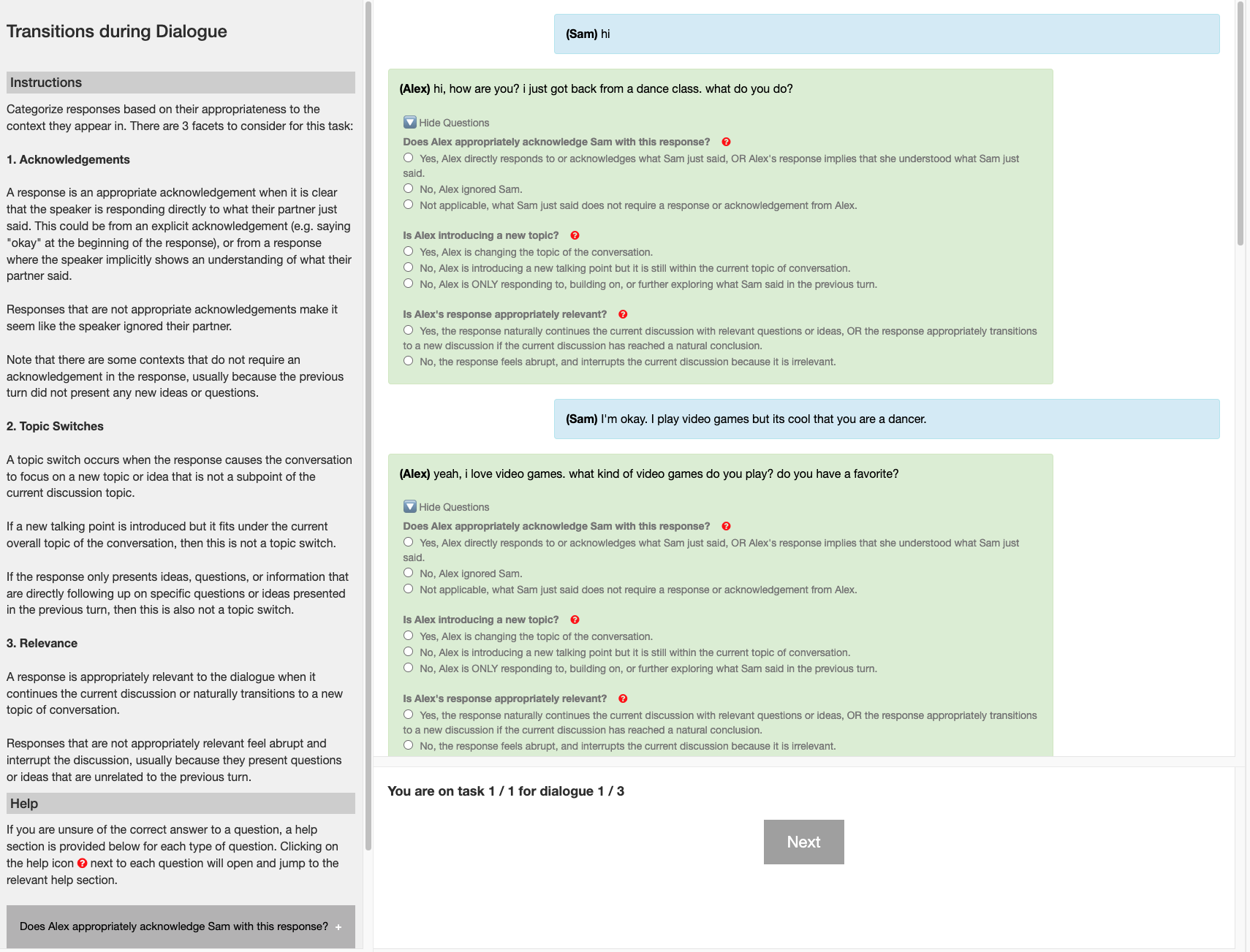}
    \caption{Interface for \texttt{ignore}, \texttt{irrelevant}, \texttt{follow-up}, and \texttt{topic switch}}
    \label{fig:interface_flow}
\end{figure*}

\begin{figure*}
    \centering
    \includegraphics[width=0.9\textwidth]{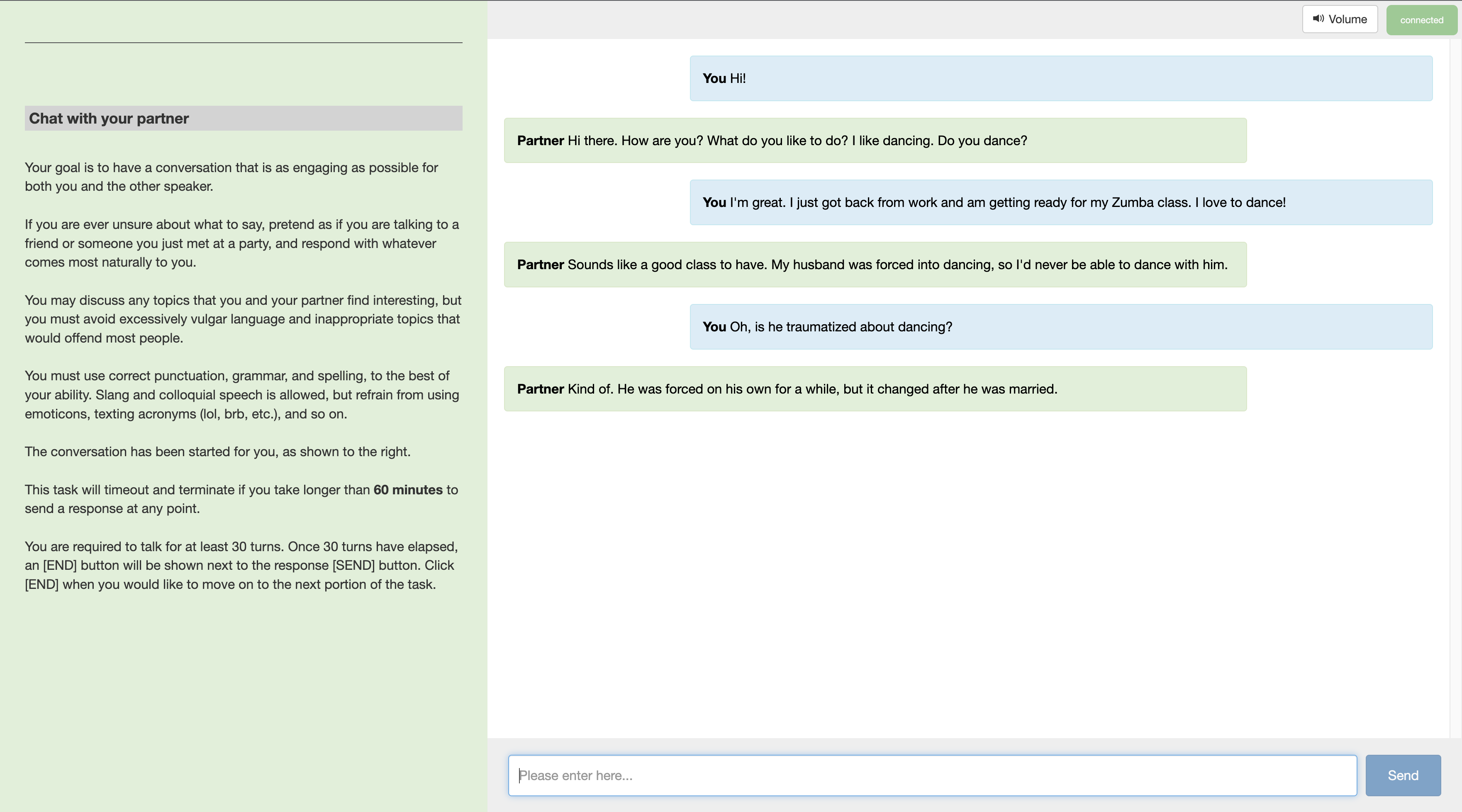}
    \caption{Interface for conversation collection}
    \label{fig:interface_conversation_collection}
\end{figure*}

Examples of the annotation interfaces for each annotation task of ABC-Eval are provided in Figures \ref{fig:interface_interpretability} - \ref{fig:interface_flow}, and an example for the conversation collection interface is provided in Figure \ref{fig:interface_conversation_collection}.

\begin{figure*}
    \centering
    \includegraphics[width=0.9\textwidth]{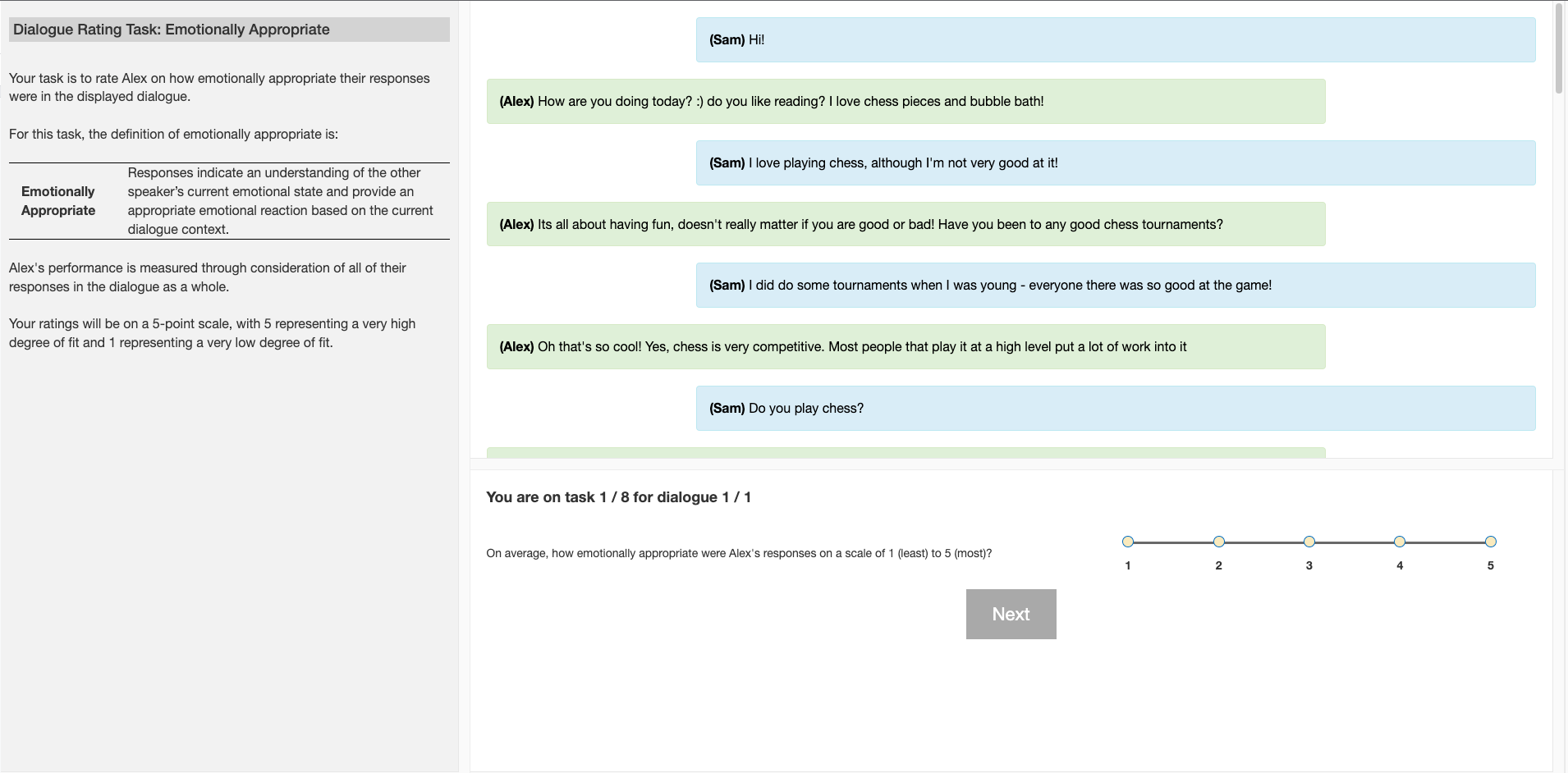}
    \caption{Interface for one dimension of Dialogue Likert}
    \label{fig:interface_dialogue_likert}
\end{figure*}

\begin{figure*}
    \centering
    \includegraphics[width=0.9\textwidth]{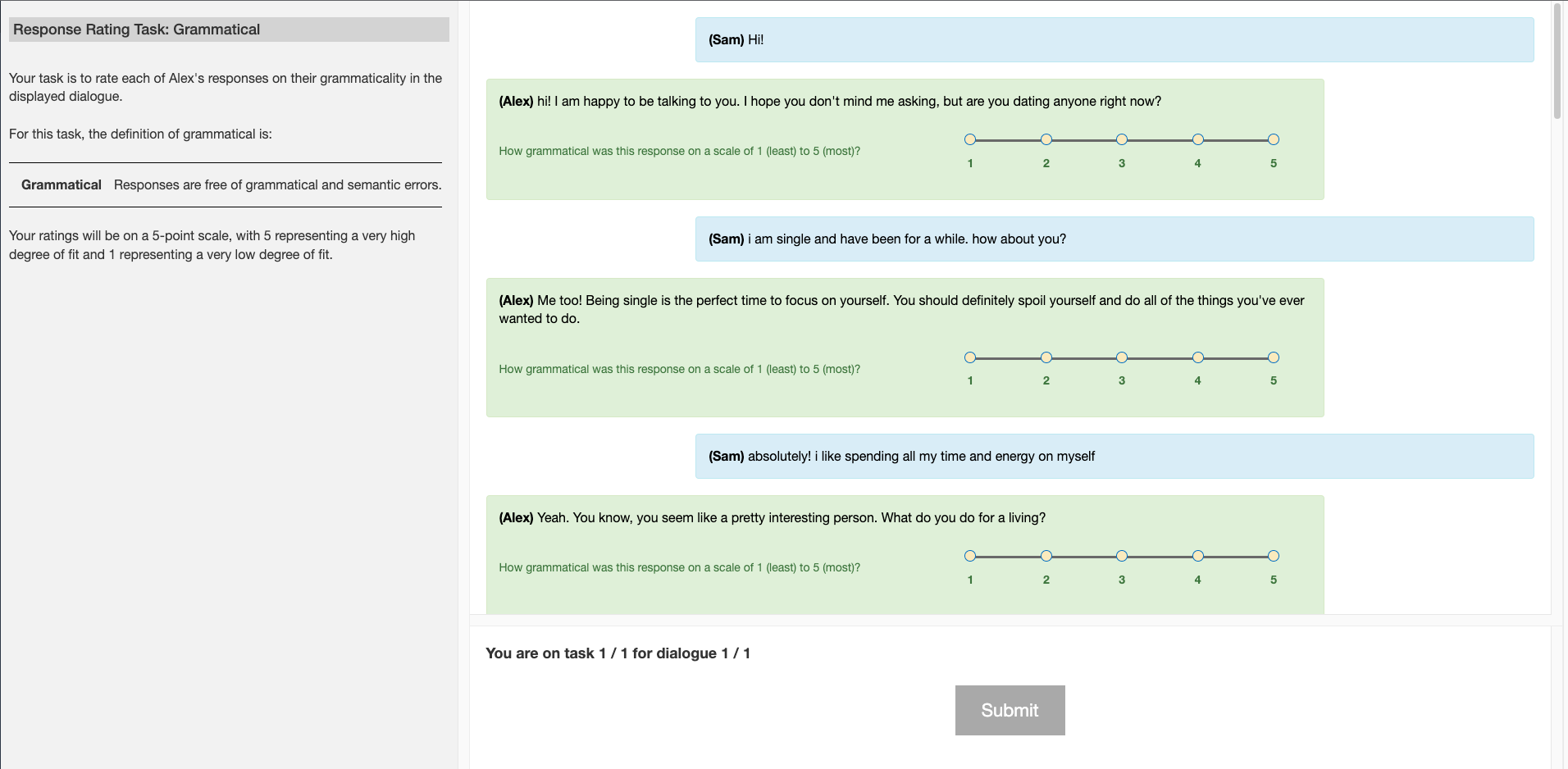}
    \caption{Interface for one dimension of Turn Likert}
    \label{fig:interface_turn_likert}
\end{figure*}

\begin{figure*}
    \centering
    \includegraphics[width=0.9\textwidth]{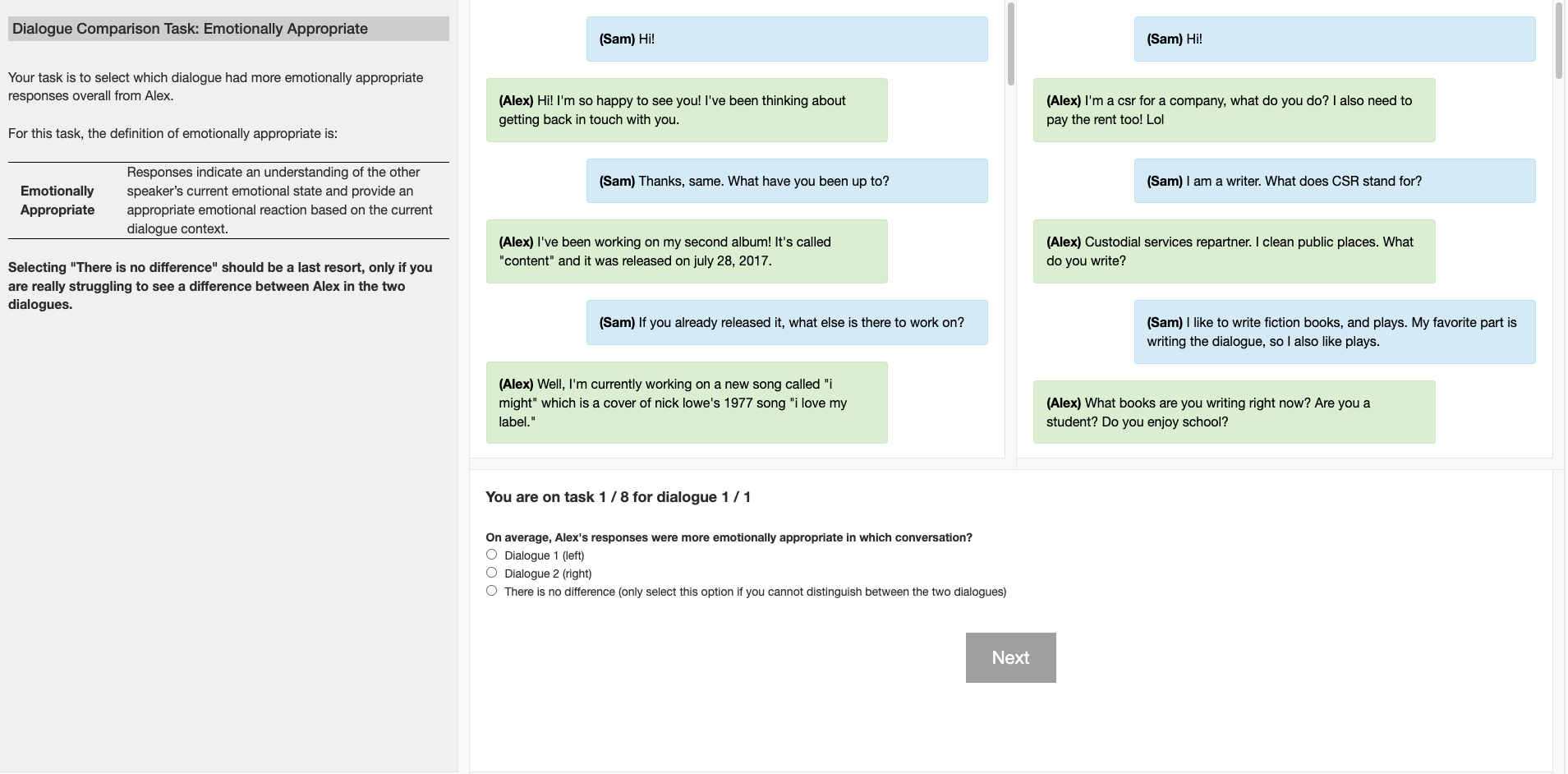}
    \caption{Interface for one dimension of Comparative}
    \label{fig:interface_comparative}
\end{figure*}

Examples of the annotation interfaces for Dialogue Likert, Turn Likert, and Comparative evaluations are provided in Figures \ref{fig:interface_dialogue_likert}, \ref{fig:interface_turn_likert}, and \ref{fig:interface_comparative}, respectively. The definitions that were shown to the annotators in the interface for each of the 8 dimensions of Dialogue Likert, Turn Likert, and Comparative are taken verbatim from \citet{finch_towards:20}.

\end{document}